\documentclass[10pt,journal,compsoc]{IEEEtran}

\ifCLASSOPTIONcompsoc
  \usepackage[nocompress]{cite}
\else
  \usepackage{cite}
\fi

\ifCLASSINFOpdf
  \usepackage[pdftex]{graphicx}
\else
\fi

\usepackage{amsmath}
\usepackage{array}

\ifCLASSOPTIONcompsoc
 \usepackage[caption=false,font=footnotesize,labelfont=sf,textfont=sf]{subfig}
\else
 \usepackage[caption=false,font=footnotesize]{subfig}
\fi

\usepackage{url}

\usepackage{amssymb}
\usepackage{booktabs}
\usepackage[table]{xcolor}
\definecolor{darkgreen}{rgb}{0.0, 0.5, 0.0}
\usepackage{colortbl}

\usepackage{makecell}

\usepackage[pagebackref=false,breaklinks=true,colorlinks=true,bookmarks=false]{hyperref}
\usepackage[normalem]{ulem}

\definecolor{robo_blue}{RGB}{66, 133, 244}
\definecolor{robo_red}{RGB}{231, 66, 52}
\definecolor{robo_yellow}{RGB}{251, 189, 5}
\definecolor{robo_green}{RGB}{51, 168, 82}
\definecolor{robo_gray}{RGB}{165, 165, 165}

\begin{document}

\title{Benchmarking and Improving Bird's Eye View Perception Robustness in Autonomous Driving}

\author{Shaoyuan~Xie,
        Lingdong~Kong,
        Wenwei~Zhang,
        Jiawei~Ren,
        Liang~Pan,
        Kai~Chen,
        and~Ziwei~Liu
\IEEEcompsocitemizethanks{
    \IEEEcompsocthanksitem S. Xie is with the Donald Bren School of Information and Computer Sciences, University of California, Irvine, CA, USA.
    \IEEEcompsocthanksitem L. Kong is with the School of Computing, National University of Singapore, and CNRS@CREATE, Singapore.
    \IEEEcompsocthanksitem J. Ren and Z. Liu are with Nanyang Technological University, Singapore.
    \IEEEcompsocthanksitem W. Zhang, L. Pan, and K. Chen are with Shanghai AI Laboratory, China.
    \IEEEcompsocthanksitem Z. Liu serves as the corresponding author. E-mail: \url{ziwei.liu@ntu.edu.sg}.
    }
}

\IEEEtitleabstractindextext{
\begin{abstract}
    Recent advancements in bird's eye view (BEV) representations have shown remarkable promise for in-vehicle 3D perception. However, while these methods have achieved impressive results on standard benchmarks, their robustness in varied conditions remains insufficiently assessed. In this study, we present RoboBEV, an extensive benchmark suite designed to evaluate the resilience of BEV algorithms. This suite incorporates a diverse set of camera corruption types, each examined over three severity levels. Our benchmarks also consider the impact of complete sensor failures that occur when using multi-modal models. Through RoboBEV, we assess 33 state-of-the-art BEV-based perception models spanning tasks like detection, map segmentation, depth estimation, and occupancy prediction. Our analyses reveal a noticeable correlation between the model's performance on in-distribution datasets and its resilience to out-of-distribution challenges. Our experimental results also underline the efficacy of strategies like pre-training and depth-free BEV transformations in enhancing robustness against out-of-distribution data. Furthermore, we observe that leveraging extensive temporal information significantly improves the model's robustness. Based on our observations, we design an effective robustness enhancement strategy based on the CLIP model. The insights from this study pave the way for the development of future BEV models that seamlessly combine accuracy with real-world robustness. The benchmark toolkit and model checkpoints are publicly accessible at: \url{https://github.com/Daniel-xsy/RoboBEV}.
\end{abstract}

}

\maketitle

\IEEEdisplaynontitleabstractindextext
\ifCLASSOPTIONpeerreview
\begin{center} \bfseries EDICS Category: 3-BBND \end{center}
\fi
\IEEEpeerreviewmaketitle

\IEEEraisesectionheading{\section{Introduction}
\label{sec:introduction}}

\IEEEPARstart{D}{eep} neural network-based 3D perception methods have registered transformative breakthroughs, excelling in a range of demanding benchmarks~\cite{li2022bevformer, wang2022detr3d, huang2021bevdet, liu2022petr, wang2022probabilistic, wang2021fcos3d, lang2019pointpillars, vora2020pointpainting, zhou2018voxelnet, yan2018second}. Among these, camera-centric methods~\cite{li2022bevformer, wang2022detr3d, huang2021bevdet, liu2022petr, wang2022probabilistic, wang2021fcos3d} have surged in popularity over their LiDAR-driven counterparts~\cite{lang2019pointpillars, vora2020pointpainting, zhou2018voxelnet, yan2018second}, primarily due to advantages such as reduced deployment costs, augmented computational efficiency, and the ability to provide dense semantic insights~\cite{ma2022vision,li2024survey}. Central to many of these advancements is the bird's eye view (BEV) representation, which offers a trio of significant benefits:
\begin{itemize}
    \item It facilitates unified learning from multi-view images.
    
    \item It encourages a physically interpretable methodology for fusing information across temporal instances.
    
    \item Its output domain aligns seamlessly with several downstream applications, such as prediction and planning, which fortifies the performance metrics of vision-centric 3D perception frameworks.
\end{itemize}

However, this blossoming landscape of BEV perception methodologies is not without its challenges. Despite their evident prowess, the resilience of these algorithms in the face of out-of-context or unforeseen scenarios remains under-examined. This oversight is particularly concerning given that many of these algorithms are envisioned to function in safety-critical realms such as autonomous driving. Traditionally, the robustness of algorithms can be bifurcated into adversarial robustness~\cite{carlini2017towards, szegedy2013intriguing,goodfellow2014explaining, moosavi2017universal, madry2017towards,brown2017adversarial, xie2017adversarial} -- which delves into worst-case scenarios -- and robustness under distribution shift~\cite{hendrycks2019benchmarking,barbu2019objectnet,recht2019imagenet,hendrycks2021natural,recht2019imagenet} that examines average-case performance, and, to certain extent, mirroring real-world conditions.

\begin{figure*}[t]
    \centering
    \includegraphics[width=\linewidth]{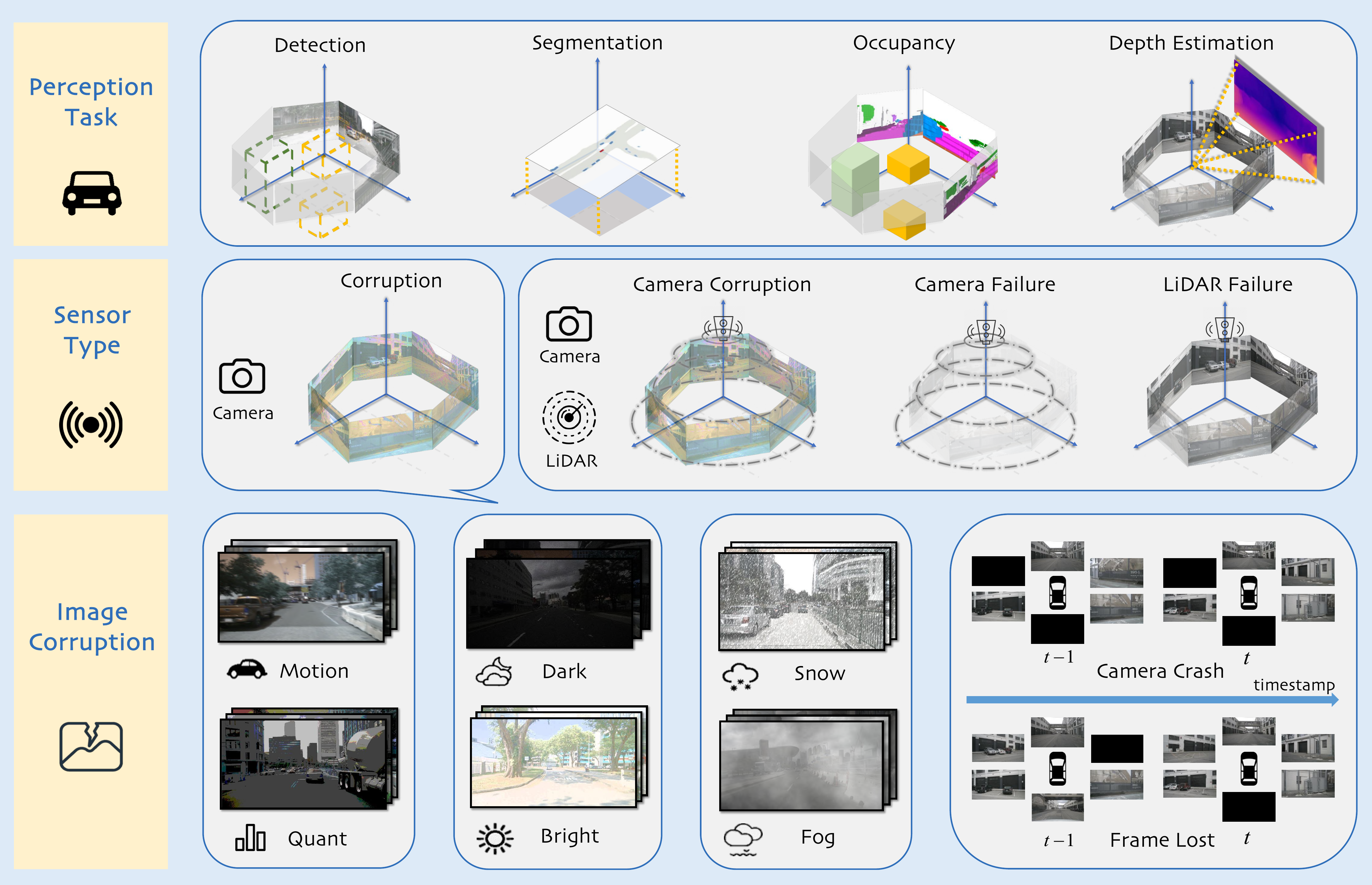}
    \vspace{-0.6cm}
    \caption{\textbf{\textit{RoboBEV} benchmark designs}. The benchmark comprehensively encompasses \textbf{four} distinct BEV perception tasks (detection, segmentation, occupancy prediction, and depth estimation), \textbf{four} diverse sensor type configurations in between LiDAR, cameras, and joint setups (camera corruption, camera failure, and LiDAR failure), and an array of \textbf{eight} natural image corruptions (Brightness, Darkness, Fog, Snow, Motion Blur, Color Quantization, Camera Crash, and Frame Lost), each categorized into \textbf{three} distinct severity levels.}
    \label{fig:taxonomy}
    \vspace{0.2cm}
\end{figure*}   

While adversarial robustness of 3D perception models has been studied \cite{xie2023adversarial, rossolini2022real, cao2021invisible, tu2020physically}, this work seeks to explore a less-traveled avenue: robustness of BEV-centric 3D perception systems when subjected to natural, often unpredictable, corruptions.
In this work, to address the existing knowledge gap, we present a comprehensive benchmark dubbed \textit{RoboBEV}. This benchmark evaluates the robustness of BEV perceptions against natural corruptions including exterior environments, interior sensors, and temporal factors. Specifically, the exterior environments include various lighting and weather conditions, which are simulated by incorporating \textit{Brightness}, \textit{Dark}, \textit{Fog}, and \textit{Snow} corruption types. Additionally, the inputs may be corrupted by interior factors caused by sensors, such as \textit{Motion Blur} and \textit{Color Quant}. We further propose two novel corruptions in consecutive space tailored for BEV-based temporal fusion strategies, namely \textit{Camera Crash} and \textit{Frame Lost}. Moreover, we consider complete sensor failure for camera-LiDAR fusion models~\cite{liu2022bevfusion, bai2022transfusion,chen2022autoalignv2} that are trained on multi-modal inputs. The study involves a comprehensive investigation of diverse out-of-distribution corruption settings that are highly relevant to real-world autonomous driving applications. Figure~\ref{fig:taxonomy} summarizes the diverse BEV perception tasks and corruption types in our benchmark study.

Leveraging the proposed \textit{RoboBEV} benchmark, we conduct an exhaustive analysis of 33 BEV perception models under corruptions across various severity levels. Finally, based on the observations, we propose to improve the model's robustness by leveraging the CLIP~\cite{radford2021learning} backbone and adapting it to BEV perception tasks. The key contributions of this work are summarized as follows:
\begin{enumerate}
    \item We introduce \textit{RoboBEV}, a comprehensive benchmark suite for evaluating BEV perception robustness under a diverse set of natural corruptions.
    
    \item We conduct extensive experiments to assess the performance of 30 camera-based and 3 camera-LiDAR fusion-based BEV perception algorithms. These algorithms are evaluated across eight distinct corruptions, each applied at three different severity levels, for a total of four perception tasks.
    
    \item Our study offers valuable insights through in-depth analyses of the factors that contribute to robustness under corruption scenarios, which shed light on future model designs. Our key observations are: i) The absolute performances show a strong correlation with the performances under corruptions. However, the relative robustness does not necessarily increase as standard performance improves; ii) The model pre-training together with depth-free BEV transformation has great potential to enhance robustness; and iii) Utilizing long and rich temporal information largely enhances the robustness.
    
    \item Based on our observations, we propose to leverage the CLIP~\cite{radford2021learning} model as the backbone to improve the robustness of BEV perception models further.
    
    \item We make the datasets and benchmark toolkit publicly available, aiming to encourage the research community to replicate and extend our findings.
\end{enumerate}

The remainder of this paper is organized as follows. Section~\ref{sec:related_works} reviews the relevant literature on vision-centric and LiDAR-based BEV perception, out-of-distribution robustness, and robustness enhancement using CLIP. Section~\ref{sec:bev_preliminaries} provides necessary preliminaries of BEV-based perception tasks. Section~\ref{sec:benchmark_design} elaborates in detail on our benchmark designs and robustness metrics. Extensive experimental studies are included in Section~\ref{sec:exp}. Based on the results, we draw analyses and observations in Section~\ref{sec:dis}. Finally, Section~\ref{sec:conclusion} discusses potential limitations and offers concluding remarks.
\section{Related Works}
\label{sec:related_works}
In this section, we review the most relevant works on the topics of BEV perception, out-of-distribution robustness, and popular robustness enhancement strategies.

\subsection{Camera-Based BEV Perception}
BEV perception methodologies can be divided into two primary branches predicated on the explicitness of their depth estimation \cite{li2024survey}. A segment of the literature, influenced by LSS~\cite{philion2020lift}, such as BEVDet~\cite{huang2021bevdet}, employs an auxiliary depth estimation branch to facilitate the transformation from perspective view to bird's eye view (PV2BEV). BEVDepth~\cite{li2022bevdepth} refines this paradigm, enhancing depth estimation accuracy using explicit depth data from point clouds. Meanwhile, BEVerse~\cite{zhang2022beverse} introduces a multi-task learning framework that achieves benchmark-setting outcomes. In contrast, an alternative research trajectory avoids explicit depth estimation. Drawing inspiration from DETR~\cite{carion2020end}, models like DETR3D~\cite{wang2022detr3d} and ORA3D~\cite{roh2022ora3d} encapsulate 3D objects as queries, leveraging Transformers' cross-attention mechanisms. Following this, PETR~\cite{liu2022petr} boosts performance by formulating 3D position-aware representations. Simultaneously, BEVFormer~\cite{li2022bevformer} and PolarFormer~\cite{jiang2022polarformer} venture into temporal cross-attention and polar coordinate-based 3D target predictions, respectively. Taking a leaf out of Sparse RCNN's~\cite{sun2021sparse} book, SRCN3D~\cite{shi2022srcn3d} and Sparse4D~\cite{lin2022sparse4d} pioneer sparse proposals for feature amalgamation. Meanwhile, SOLOFusion~\cite{Park2022TimeWT} pursues deeper historical data integration for temporal modeling. In addition to detection, BEV perception tasks also include map segmentation~\cite{bartoccioni2023lara, zhou2022cross}, multi-view depth estimation~\cite{wei2023surrounddepth}, and semantic occupancy prediction~\cite{huang2023tri, wei2023surroundocc, tong2023scene, wang2023openoccupancy, min2023occ}. While these methodologies flaunt impressive outcomes on pristine datasets, their resilience against natural corruptions remains an enigma.

\subsection{LiDAR-Based 3D Perception}
LiDAR, with its precision in capturing spatial relationships using laser beams, has paved the way for breakthroughs in 3D perception, central to applications like autonomous driving \cite{meng2022towards}. Two primary tasks have gained prominence: 3D object detection and LiDAR semantic segmentation, both of which have inherent connections to BEV perception \cite{hong2024dsnet,jaritz2023cross,kong2024lasermix2,hu2022randla,xu2023frnet}. In the realm of \textit{3D object detection}, the focus has been on optimally representing LiDAR point cloud data \cite{kong2023laserMix}. Point-based approaches, such as those presented in \cite{pointrcnn,pointgnn,std,3dssd}, shine in preserving the innate geometry of point clouds, capturing local structures and patterns. Meanwhile, voxel-based strategies, like \cite{second,voxelnet,centerpoint}, convert the irregular point clouds into structured grids, relying on sparse convolution techniques \cite{second} to handle non-empty voxels efficiently. Pillar-based techniques, highlighted by works like \cite{lang2019pointpillars,pillarnet}, offer a trade-off between detection accuracy and computational speed by fine-tuning the vertical resolution. Additionally, hybrid approaches, such as \cite{shi2022pv,pvrcnn}, merge the strengths of both point and voxel representations to derive more enriched features. On the other hand, \textit{semantic segmentation} techniques often pivot on the representation choice. Raw point methods, like \cite{thomas2019kpconv,puy2023waffleiron}, emphasize the direct usage of irregular point clouds, while range view approaches, showcased in \cite{wu2018squeezeseg,milioto2019rangenet++,ando2023rangevit,kong2023conDA,kong2023rethinking}, convert these point clouds into 2D grids. This conversion aligns closely with BEV perception, transforming 3D data into a top-down perspective, which is essential for many applications. Further refining this idea are bird's eye view techniques, exemplified by \cite{zhang2020polarnet}, which offer a direct 2D top-down representation. Voxel-centric methods, such as \cite{2019Minkowski}, maintain the 3D spatial structure, often outperforming other singular modalities. Modern research, like \cite{2020AMVNet,tang2020searching,xu2021rpvnet,chen2023clip2Scene,liu2023uniseg,chen2023towards,liu2023segment,liu2024m3net}, pushes the boundaries by exploring the fusion of multiple views, seeking to harness the complementary strengths of different representations. In essence, while LiDAR-based 3D perception methodologies, especially those linked with BEV perception, have exhibited significant promise, their resilience in real-world conditions warrants deeper exploration and validation.

\subsection{Robustness under Adversarial Attacks}
Modern neural networks, while showcasing staggering capabilities, remain vulnerable to adversarial onslaughts, where meticulously engineered perturbations in inputs can precipitate erroneous outputs~\cite{szegedy2013intriguing,goodfellow2014explaining, moosavi2017universal}. The menace of adversarial examples has been a research epicenter across various vision domains: classification~\cite{szegedy2013intriguing,goodfellow2014explaining}, detection~\cite{xie2017adversarial,liu2018dpatch}, and segmentation~\cite{xie2017adversarial,rossolini2022real}. These adversarial stimuli can emerge in both digital domains~\cite{szegedy2013intriguing,goodfellow2014explaining} and real-world environments~\cite{rossolini2022real,kurakin2018adversarial}. Alarming findings reveal that adversarial examples can cripple 3D perception systems, flagging potential safety concerns during practical deployments~\cite{rossolini2022real, cao2021invisible, tu2020physically}. While Xie \textit{et al.}~\cite{xie2023adversarial} delve into the adversarial robustness of camera-centric detectors, our focus pivots towards more pervasive natural corruptions.

\subsection{Robustness under Natural Corruptions}
Assessing model tenacity against corruptions has burgeoned as a pivotal research domain. Several benchmarks, such as ImageNet-C~\cite{hendrycks2019benchmarking}, ObjectNet~\cite{barbu2019objectnet}, ImageNetV2~\cite{recht2019imagenet}, and more, evaluate the robustness of 2D image classifiers against an array of corruptions. For instance, ImageNet-C taints pristine ImageNet samples with simulated anomalies like compression artifacts and motion blurs. On the other hand, ObjectNet~\cite{barbu2019objectnet} offers a test set abundant in rotation, background, and viewpoint variances. Hendrycks \textit{et al.}~\cite{hendrycks2021many} underscore the correlation between synthetic corruption robustness and enhancements in real-world scenarios. Recently, Some works~\cite{ge2023metabev,man2023dualcross,kong2023robo3d} endeavor to improve the robustness of 3D perception models. Kong \textit{et al.} \cite{kong2023robodepth,kong2023robodepth_challenge} establish a robustness benchmark for monocular depth estimation under corruptions. Ren \textit{et al.} \cite{ren2022modelnetc} design atomic corruptions on indoor object-centric point clouds and CAD models to understand classifiers' robustness.
Yet, a void persists concerning benchmarks for 3D BEV perception models, which play critical roles in safety-sensitive applications. While a concurrent study by Zhu \textit{et al.}~\cite{zhu2023understanding} explores a similar landscape, their narrative is predominantly adversarial-centric. In contrast, our benchmarks, spanning models, tasks, scenarios, and validation studies, offer a broader and more comprehensive lens into this domain.

\begin{figure*}[t]
    \centering
    \includegraphics[width=\linewidth]{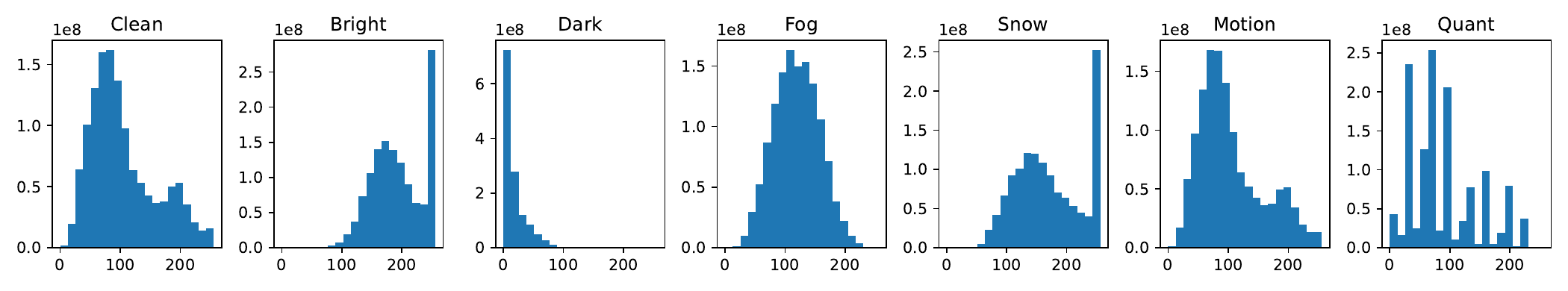}
    \vspace{-0.6cm}
    \caption{Histograms of pixel distributions for different corruption types. While certain corruptions exhibit minimal shifts in pixel distribution (\textit{e.g.}, Motion Blur), it is noteworthy that these alterations predominantly have adverse effects on the overall performance of the BEV perception systems.}
    \label{fig:hist}
\end{figure*}

\subsection{Robustness Enhancements using CLIP}
The Contrastive Language-Image Pre-training (CLIP) model \cite{radford2021learning} has been shown to significantly improve the model's out-of-distribution robustness compared to previous supervised trained models on ImageNet\cite{miller2021accuracy, nguyen2022quality}. Recent studies have begun to explore why CLIP exhibits superior robustness~\cite{nguyen2022quality, xue2023understanding, fang2022data} and how this robustness can be maintained after fine-tuning~\cite{wortsman2022model, goyal2023finetune}. Specifically, \cite{goyal2023finetune} found that while end-to-end fine-tuning can enhance in-distribution performance on supervised datasets, it can also compromise the out-of-distribution robustness of the pre-trained CLIP model. To address this, they employed weighted parameter adjustments to achieve both performance and robustness. Given CLIP's robust performance, a natural question arises: can we leverage the pre-trained CLIP model to enhance BEV perception robustness? In this work, we make the first attempt to investigate how the robustness of the pre-trained CLIP model can be retained.
\section{BEV Perception Preliminaries} 
\label{sec:bev_preliminaries}
In this section, we delineate the commonly adopted techniques for BEV perception algorithms, which have demonstrated enhanced performance across standard datasets.

\subsection{Model Pre-Training}
Over the past few years, pre-training has enhanced the performance of computer vision models in diverse tasks. Within the domain of camera-driven 3D perception, initializing the ResNet backbone using FCOS3D~\cite{wang2021fcos3d} weights has become standard practice. To stabilize the training process, FCOS3D adjusts a depth weight from 0.2 to 1 during fine-tuning~\cite{wang2021fcos3d}. Another prevailing approach involves training the VoVNet-V2~\cite{lee2020centermask} backbone on the DDAD15M~\cite{guizilini20203d} dataset, targeting depth estimation, before fine-tuning it using the nuScenes training set for detection. Semantically, these pre-training techniques fall into two categories: semantic and depth pre-training. Furthermore, M-BEV~\cite{chen2024m} introduces robust masked image pre-training techniques designed to augment model resilience in scenarios characterized by absent sensor data.

\subsection{Temporal Fusion}
The dynamic landscape of autonomous driving demands precise velocity estimates of moving entities, which is a challenge for singular frame inputs. This reveals the importance of temporal cues in enhancing vision systems' perception capabilities. Prior research has pioneered various methodologies to harness these temporal cues. For instance, BEVFormer~\cite{li2022bevformer} integrates history data and leverages temporal cross-attention to aggregate BEV features from multi-timestamp images. Meanwhile, BEVDet4D~\cite{huang2022bevdet4d} appends features from multiple frames to weave in temporal nuances, and SOLOFusion~\cite{Park2022TimeWT} aims for more inclusive temporal modeling by merging extensive historical data. However, the resilience of these sophisticated temporal models under corrupted conditions remains largely under-explored.

\subsection{Camera-LiDAR Fusion}
The BEV paradigm streamlines the fusion of features from a variety of input modalities. While some algorithms focus on crafting BEV representations solely from images, a notable fraction of the literature, including works like \cite{liu2022bevfusion, liang2022bevfusion, chen2022autoalign, chen2022autoalignv2, han2023exploring, chen2023futr3d}, advocates for a unified BEV space. This aligns features extracted from both images and point clouds. We delve deep into the performance of such multi-modal fusion algorithms, especially under circumstances where images are corrupted, yet the LiDAR mechanism remains untouched. Furthermore, we address a common scenario where the model is trained using multi-modal inputs but deployed on vehicles equipped with only one of the sensors. To assess robustness, we evaluate the model's performance under conditions of complete sensor failure, where either the camera or LiDAR is missing.

\subsection{BEV View Transformation}
The body of works in BEV transformation can be divided based on the use of depth estimation techniques \cite{li2024survey}. One faction, as discussed in \cite{huang2021bevdet, li2022bevdepth, zhang2022beverse, Park2022TimeWT}, embeds a distinct depth-estimation branch within their systems. Given the inherent challenges in predicting 3D bounding boxes from images, these models first forecast a per-pixel depth map. This map then serves as a compass, guiding image features to their rightful 3D coordinates. The subsequent BEV transformation process often follows a bottom-up approach, as depicted in \cite{wang2022detr3d}. On the other side of the spectrum are models leveraging pre-defined object queries \cite{li2022bevformer, wang2022detr3d} or lean proposals \cite{lin2022sparse4d, shi2022srcn3d} to collate 2D features in a top-down manner. While both these paradigms have demonstrated their prowess on original datasets, we further expand the horizon by examining their efficacy on corrupted data.

\begin{table*}[t]
    \centering
    \caption{Severity level setups in corruption simulations: Detailed parameters used for generating multi-level corruptions for each corruption type.}
    \vspace{-0.2cm}
    \begin{tabular}{c|c|c|c|c}
        \toprule
        \textbf{Corruption} & \textbf{Parameter} & \textbf{Easy} & \textbf{Moderate} & \textbf{Hard} 
        \\\midrule\midrule
        \rowcolor{gray!10}Bright & adjustment in HSV space & 0.2 & 0.4 & 0.5 
        \\
        Dark & scale factor & 0.5 & 0.4 & 0.3 
        \\
        \rowcolor{gray!10}Fog & (thickness, smoothness) & (2.0, 2.0) & (2.5, 1.5) & (3.0, 1.4)
        \\
        Snow & \makecell{(mean, std, scale, threshold, \\blur radius, blur std, blending ratio)}	 & (0.1, 0.3, 3.0, 0.5, 10.0, 4.0, 0.8)	 & (0.2, 0.3, 2, 0.5, 12, 4, 0.7) & (0.55, 0.3, 4, 0.9, 12, 8, 0.7)
        \\
        \rowcolor{gray!10}Motion & (radius, sigma) & (15, 5) & (15, 12) & (20, 15) 
        \\
        Quant & bit number & 5 & 4 & 3
        \\
        \rowcolor{gray!10}Crash & number of dropped camera & 2 & 4 & 5
        \\
        Frame & probability of frame dropping  & 2/6 & 4/6 & 5/6
        \\
        \bottomrule
    \end{tabular}
    \label{tab:my_label}
\end{table*}
\section{Benchmark Design}
\label{sec:benchmark_design}
This section elaborates on our benchmark design for BEV algorithms. Subsections~\ref{sec:dataset} and \ref{sec:corruption} discuss the creation and structure of the \textit{nuScenes-C} benchmark dataset. Subsection~\ref{sec:metric} illustrates the robustness metrics designed to assess the resilience of models against various corruptions.

\subsection{Dataset Generation}
\label{sec:dataset}
Our main proposal is the \textit{nuScenes-C} benchmark dataset, which is created by introducing corruptions to the validation set of the nuScenes dataset~\cite{caesar2020nuscenes}. Encompassing a vast expanse of eight distinct corruptions, our dataset simulates challenges posed by external environmental elements, sensor-induced distortions, and our innovative temporal corruptions.
Inspired by \cite{hendrycks2019benchmarking}, we tier each corruption type across three intensities: easy, moderate, and hard. These severity levels ensure that while challenges are present, they do not entirely destroy performance, thereby maintaining the relevance and integrity of our findings. Moreover, we introduce variability within each severity to ensure diversity. Comprehensively, our benchmark consists of 866,736 images, each with a resolution of 1600$\times$900 pixels.

We also consider scenarios simulating complete sensor blackouts in our camera-LiDAR fusion algorithms. Every pixel is set to zero when simulating the camera's absence. To emulate the missing LiDAR readings, only the data points within a $[-$45, 45$]$ degree frontal field-of-view (FOV) are retained. Such a design choice is rooted in our observations that multi-modal trained models suffer from significant performance drops when LiDAR readings are entirely absent.

\subsection{Natural Corruptions}
\label{sec:corruption}
A visual guide to our corruption taxonomy is presented in Figure~\ref{fig:taxonomy}. Broadly, we focus on three corruption categories. First, those induced by external environmental dynamics, such as varying illumination or meteorological extremes, are simulated via \textit{Brightness}, \textit{Dark}, \textit{Fog}, and \textit{Snow}. Considering the bulk of training data is captured under relatively benign conditions, testing models under these extremes is crucial.

Secondly, sensor-driven distortions can corrupt collected imagery. High-speed motion may induce blur, or memory conservation tactics might compel image quantization. To mimic these real-world challenges, we have integrated \textit{Motion Blur} and \textit{Color Quant}.

Lastly, we introduce camera malfunctions, where entire image sets or random frames are omitted due to hardware issues. This is captured by our novel \textit{Camera Crash} and \textit{Frame Lost} corruptions. The illustration of these processes is visualized in Figure~\ref{fig:taxonomy}. We visualize the pixel histogram analysis on our synthesized images, as shown in Figure~\ref{fig:hist}. A notable observation is that the \textit{Motion Blur} corruption, while inducing minimal pixel distribution shifts, still caused a significant performance dip. Additional experimental findings and results are discussed in detail in Section~\ref{sec:exp}.

\begin{table*}[t]
    \centering
    \caption{BEV model calibration. Pretrain: model initialized from pretrained FCOS3D~\cite{wang2021fcos3d} checkpoint; Temporal: model utilizes temporal information; Depth: model with explicit depth estimation branch used in the pipeline; CBGS: model uses the class-balanced group-sampling training strategy~\cite{zhu2019class}. The mCE and mRR scores are given in percentage (\%). Bold: Best in the category. \underline{Underline}: Second best in the category.}
    \label{tab:robodet_model}
    \vspace{-0.2cm}
    \scalebox{0.95}{
    \begin{tabular}{r|cccc|c|c|c|ccc}
    \toprule
    \textbf{Model} & \textbf{Pretrain} & \textbf{Temporal} & \textbf{Depth} & \textbf{CBGS} &\textbf{Backbone} & \textbf{BEV Encoder} & \textbf{Image Size} & \textbf{NDS} $\uparrow$ & \textbf{mCE} $\downarrow$ & \textbf{mRR} $\uparrow$\\
    \midrule\midrule
    \rowcolor{gray!10}DETR3D~\cite{wang2022detr3d} & \checkmark &  &  &  & ResNet & Attention & 1600 $\times$ 900 & 0.4224 & 100.00 & 70.77 \\
    DETR3D$_{\text{CBGS}}$~\cite{wang2022detr3d}  & \checkmark &  &  & \checkmark & ResNet & Attention & 1600 $\times$ 900 & 0.4341 & 99.21 & 70.02   
    \\
    \rowcolor{gray!10}BEVFormer~{\scriptsize (small)}~\cite{li2022bevformer}  & \checkmark & \checkmark &  &  & ResNet & Attention & 1280 $\times$ 720 & \underline{0.4787}  & 101.23 & 59.07  
    \\
    BEVFormer-S~{\scriptsize (small)}~\cite{li2022bevformer}   & \checkmark &  &  &  & ResNet & Attention & 1280 $\times$ 720 & 0.2622 & 114.43 & \textbf{76.87}  
    \\
    \rowcolor{gray!10}BEVFormer~{\scriptsize (base)}~\cite{li2022bevformer}  & \checkmark & \checkmark &  &  & ResNet & Attention & 1600 $\times$ 900 & \textbf{0.5174} & \underline{97.97} & 60.40  
    \\
    BEVFormer-S~{\scriptsize (base)}~\cite{li2022bevformer} & \checkmark &  &  &  & ResNet & Attention & 1600 $\times$ 900 & 0.4129 & 101.87 & 69.33  
    \\
    \rowcolor{gray!10}PETR~{\scriptsize (r50)}~\cite{liu2022petr}  &  &  &  &  & ResNet & Attention & 1408 $\times$ 512 & 0.3665 & 111.01 & 61.26  \\
    PETR~{\scriptsize (vov)}~\cite{liu2022petr}  & \checkmark &  &  &  & VoVNet-V2 & Attention & 1600 $\times$ 640 & 0.4550 & 100.69 & 65.03   
    \\
    \rowcolor{gray!10}ORA3D~\cite{roh2022ora3d}  & \checkmark &  &  &  & ResNet & Attention & 1600 $\times$ 900 & 0.4436 & 99.17 & 68.63  
    \\
    PolarFormer~{\scriptsize (r101)}~\cite{jiang2022polarformer}  & \checkmark &  &  &  & ResNet & Attention & 1600 $\times$ 900 & 0.4602 & \textbf{96.06} & \underline{70.88}   \\
    \rowcolor{gray!10}PolarFormer~{\scriptsize (vov)}~\cite{jiang2022polarformer}    & \checkmark &  &  &  & VoVNet-V2 & Attention & 1600 $\times$ 900 & 0.4558 & 98.75 & 67.51  
    \\
    \midrule
    SRCN3D~{\scriptsize (r101)}~\cite{shi2022srcn3d}  & \checkmark &  &  &  & ResNet & CNN + Attn. & 1600 $\times$ 900 & 0.4286 & \textbf{99.67} & \textbf{70.23}   
    \\
    \rowcolor{gray!10}SRCN3D~{\scriptsize (vov)}~\cite{shi2022srcn3d}  & \checkmark &  &  &  & VoVNet-V2 & CNN + Attn. & 1600 $\times$ 900 & 0.4205 & 102.04 & 67.95
    \\
    Sparse4D~{\scriptsize (r101)}~\cite{lin2022sparse4d}  & \checkmark & \checkmark &  &  & ResNet & CNN + Attn. & 1600 $\times$ 640 & \textbf{0.5438}  & 100.01 & 55.04   \\
    \midrule
    \rowcolor{gray!10}BEVDet~{\scriptsize (r50)}~\cite{huang2021bevdet}  &  &  & \checkmark & \checkmark & ResNet & CNN & 704 $\times$ 256 & 0.3770 & 115.12 & 51.83   
    \\
    BEVDet~{\scriptsize (r101)}~\cite{huang2021bevdet}  &  &  & \checkmark & \checkmark & ResNet & CNN & 704 $\times$ 256 & 0.3877 & 113.68 & 53.12    
    \\
    \rowcolor{gray!10}BEVDet~{\scriptsize (r101)}~\cite{huang2021bevdet}  & \checkmark &  & \checkmark & \checkmark & ResNet & CNN & 704 $\times$ 256 & 0.3780 & 112.80 & 56.35    
    \\
    BEVDet~{\scriptsize (tiny)}~\cite{huang2021bevdet}  &  &  & \checkmark & \checkmark & SwinTrans & CNN & 704 $\times$ 256 & 0.4037 & 116.48 & 46.26   
    \\
    \rowcolor{gray!10}BEVDepth~{\scriptsize (r50)}~\cite{li2022bevdepth}  &  &  & \checkmark & \checkmark & ResNet & CNN & 704 $\times$ 256 & 0.4058 & 110.02 & 56.82  
    \\
    BEVerse~{\scriptsize (swin-t)}~\cite{zhang2022beverse}  &  & \checkmark & \checkmark & \checkmark & SwinTrans & CNN & 704 $\times$ 256 & 0.4665 & 110.67 & 48.60  \\
    \rowcolor{gray!10}BEVerse-S~{\scriptsize (swin-t)}~\cite{zhang2022beverse} &  &  & \checkmark & \checkmark & SwinTrans & CNN & 704 $\times$ 256 & 0.1603 & 137.25 & 28.24 \\
    BEVerse~{\scriptsize (swin-s)}~\cite{zhang2022beverse}  &  & \checkmark & \checkmark & \checkmark & SwinTrans & CNN & 1408 $\times$ 512 & \underline{0.4951} & 117.82 & 49.57   
    \\
    \rowcolor{gray!10}BEVerse-S~{\scriptsize (swin-s)}~\cite{zhang2022beverse} &  &  & \checkmark & \checkmark & SwinTrans & CNN & 1408 $\times$ 512 & 0.2682 & 132.13 & 29.54 
    \\
    SOLOFusion~{\scriptsize (short)}~\cite{Park2022TimeWT} &  & \checkmark & \checkmark &  & ResNet & CNN & 704 $\times$ 256 & 0.3907 & 108.68 & 61.45 
    \\
    \rowcolor{gray!10}SOLOFusion~{\scriptsize (long)}~\cite{Park2022TimeWT} &  & \checkmark & \checkmark &  & ResNet & CNN & 704 $\times$ 256 & 0.4850 & \underline{97.99} & \underline{64.42} 
    \\
    SOLOFusion~{\scriptsize (fusion)}~\cite{Park2022TimeWT} &  & \checkmark & \checkmark & \checkmark & ResNet & CNN & 704 $\times$ 256 & \textbf{0.5381} & \textbf{92.86} & \textbf{64.53} 
    \\
    \bottomrule
    \end{tabular}
    }
\end{table*}

\begin{table*}[t]
    \centering
    \caption{The Corruption Error (CE) of each BEV detector in our \textit{RoboBEV} benchmark. The CE and mCE scores are given in percentage (\%). Bold: Best in the category. \underline{Underlined}: Best in the row if improved upon baseline. \dag: distinguish pre-training version BEVDet.}
    \vspace{-0.2cm}
    \label{tab:robodet_ce}
    \scalebox{0.98}{
    \footnotesize
    \begin{tabular}{r|p{1.1cm}<{\centering}|p{1.1cm}<{\centering}|p{1.cm}<{\centering}p{1.cm}<{\centering}p{1.cm}<{\centering}p{1.cm}<{\centering}p{1.cm}<{\centering}p{1.cm}<{\centering}p{1.cm}<{\centering}p{1.cm}<{\centering}}
    \toprule
    \textbf{Model} & \textbf{NDS} $\uparrow$ & \textbf{mCE} $\downarrow$ & \textbf{Camera} & \textbf{Frame} & \textbf{Quant} & \textbf{Motion} & \textbf{Bright} & \textbf{Dark} & \textbf{Fog} & \textbf{Snow} 
    \\\midrule\midrule
    \rowcolor{gray!10}DETR3D~\cite{wang2022detr3d} & 0.4224 & 100.00 & 100.00 & 100.00 & 100.00 & 100.00 & 100.00 & 100.00 & 100.00 & 100.00 
    \\\midrule
    DETR3D$_{\text{CBGS}}$~\cite{wang2022detr3d} & 0.4341  & 99.21  & 98.15 & 98.90 & 99.15 & 101.62 & \underline{97.47} & \textbf{100.28} & 98.23 & 99.85 
    \\
    \rowcolor{gray!10}BEVFormer~{\scriptsize (small)}~\cite{li2022bevformer} & 0.4787  & 102.40  & 101.23 & 101.96 & \underline{98.56} & 101.24 & 104.35 & 105.17 & 105.40 & 101.29  
    \\
    BEVFormer~{\scriptsize (base)}~\cite{li2022bevformer} &\textbf{0.5174}  &  97.97 & \textbf{95.87}  & \underline{\textbf{94.42}} & \textbf{95.13} & 99.54 & 96.97 & 103.76 & 97.42 & 100.69 \\
    \rowcolor{gray!10}PETR~{\scriptsize (r50)}~\cite{liu2022petr} & 0.3665  & 111.01  & 107.55 & 105.92 & 110.33 & 104.93 & 119.36 & 116.84 & 117.02 & 106.13 \\
    PETR~{\scriptsize (vov)}~\cite{liu2022petr} & 0.4550  & 100.69  & 99.09 & 97.46 & 103.06 & 102.33 & 102.40 & 106.67 & 103.43 & \underline{\textbf{91.11}} \\
    \rowcolor{gray!10}ORA3D~\cite{roh2022ora3d} & 0.4436 & 99.17 & \underline{97.26} & 98.03 & 97.32 & 100.19 & 98.78 & 102.40 & 99.23 & 100.19 \\
    PolarFormer~{\scriptsize (r101)}~\cite{jiang2022polarformer} & 0.4602 & \textbf{96.06}  & 96.16 & 97.24 & \textbf{95.13} & \underline{\textbf{92.37}} & \textbf{94.96} & 103.22 & \textbf{94.25} & 95.17 \\
    \rowcolor{gray!10}PolarFormer~{\scriptsize (vov)}~\cite{jiang2022polarformer} & 0.4558 & 98.75  & 96.13 & 97.20 & 101.48 & 104.32 & 95.37 & 104.78 & 97.55 & \underline{93.14}  \\
    \midrule
    SRCN3D~{\scriptsize (r101)}~\cite{shi2022srcn3d} & 0.4286 & \textbf{99.67}  & \textbf{98.77} & \textbf{98.96} & \underline{\textbf{97.93}} & \textbf{100.71} & \textbf{98.80} & 102.72 & \textbf{99.54} & 99.91 \\
    \rowcolor{gray!10}SRCN3D~{\scriptsize (vov)}~\cite{shi2022srcn3d}  & 0.4205 & 102.04  & 99.78 & 100.34 & 105.13 & 107.06 & 101.93 & 107.10 & 102.27 & \underline{\textbf{92.75}} \\
    Sparse4D~{\scriptsize (r101)}~\cite{lin2022sparse4d} & \textbf{0.5438} & 100.01  & 99.80 & 99.91 & 98.05 & 102.00 & 100.30 & 103.83 & 100.46 & \underline{95.72} \\
    \midrule
    \rowcolor{gray!10}BEVDet~{\scriptsize (r50)}~\cite{huang2021bevdet} & 0.3770  & 115.12  & 105.22 & 109.19 & 111.27 & 108.18 & 123.96 & 123.34 & 123.83 & 115.93 \\
    BEVDet~{\scriptsize (tiny)}~\cite{huang2021bevdet} & 0.4037  & 116.48  & 103.50 & 106.61 & 113.18 & 107.26 & 130.19 & 131.83 & 124.01 & 115.25 \\
    \rowcolor{gray!10}BEVDet~{\scriptsize (r101)}~\cite{huang2021bevdet} & 0.3877  & 113.68  & 103.32 & 107.29 & 109.25 & 105.40 & 124.14 & 123.12 & 123.28 & 113.64 \\
    BEVDet~{\scriptsize (r101\dag)}~\cite{huang2021bevdet} & 0.3780 & 112.80  & 105.84 & 108.68 & 101.99 & 100.97 & 123.39 & 119.31 & 130.21 & 112.04 \\
    \rowcolor{gray!10}BEVDepth~{\scriptsize (r50)}~\cite{li2022bevdepth} & 0.4058  & 110.02  & 103.09 & 106.26 & 106.24 & 102.02 & 118.72 & 114.26 & 116.57 & 112.98 \\
    BEVerse~{\scriptsize (swin-t)}\cite{zhang2022beverse} & 0.4665 & 110.67  & 95.49 & \underline{94.15} & 108.46 & 100.19 & 122.44 & 130.40 & 118.58 & 115.69 \\
    \rowcolor{gray!10}BEVerse~{\scriptsize (swin-s)}~\cite{zhang2022beverse} & 0.4951 & 107.82  & \underline{92.93} & 101.61 & 105.42 & 100.40 & 110.14 & 123.12 & 117.46 & 111.48 \\
    SOLOFusion~{\scriptsize (short)}~\cite{Park2022TimeWT} & 0.3907 & 108.68  & 104.45 & 105.53 & 105.47 & 100.79 & 117.27 & 110.44 & 115.01 & 110.47 \\
    \rowcolor{gray!10}SOLOFusion~{\scriptsize (long)}~\cite{Park2022TimeWT} & 0.4850 & 97.99  & 95.80 & 101.54 & 93.83 & \underline{89.11} & 100.00 & \textbf{99.61} & 98.70 & \textbf{105.35} \\
    SOLOFusion~{\scriptsize (fusion)}~\cite{Park2022TimeWT} & \textbf{0.5381} & \textbf{92.86}  & \textbf{86.74} & \textbf{88.37} & \textbf{87.09} & \underline{\textbf{86.63}} & \textbf{94.55} & 102.22 & \textbf{90.67} & 106.64 \\
    \bottomrule
    \end{tabular}
    }
\end{table*}

\subsection{Robustness Metrics}
\label{sec:metric}
We follow the official nuScenes metric~\cite{caesar2020nuscenes} to calculate robustness metrics on the \textit{nuScenes-C} dataset. We report nuScenes Detection Score (NDS) and mean Average Precision (mAP), along with mean Average Translation Error (mATE), mean Average Scale Error (mASE), mean Average Orientation Error (mAOE), mean Average Velocity Error (mAVE) and mean Average Attribute Error (mAAE).

To better compare the robustness among different BEV detectors, we introduce two new metrics inspired by \cite{hendrycks2019benchmarking} based on NDS. The first metric is the mean corruption error (mCE), which is applied to measure the \textit{relative robustness} of candidate models compared to the baseline model:
\begin{equation}
    \text{CE}_i=\frac{\sum^{3}_{l=1}(1 - \text{NDS})_{i,l}}{\sum^{3}_{l=1}(1 - \text{NDS}_{i,l}^{\text{baseline}})}~,~~~
    \text{mCE}=\frac{1}{N}\sum^N_{i=1}\text{CE}_i~,
\end{equation}
where $i$ denotes the corruption type and $l$ is the severity level; $N$ denotes the number of corruption types in our benchmark. It should be noted that one can choose different baseline models. In this work, we resort to DETR3D~\cite{wang2022detr3d} as it offers a seminar BEV detection performance. To compare the \textit{performance discrepancy} between \textit{nuScenes-C} and the standard nuScenes dataset \cite{caesar2020nuscenes}, we define a simple mean resilience rate (mRR) metric, which is calculated across three severity levels as follows:
\begin{equation}
    \text{RR}_i=\frac{\sum^{3}_{l=1}\text{NDS}_{i,l}}{3\times \text{NDS}_{\text{clean}}}~,~~~
  \text{mRR}=\frac{1}{N}\sum^N_{i=1}\text{RR}_i~.
\end{equation}

\noindent
In our benchmark, we report both metrics for each BEV model and compare the robustness based on them.
\section{Benchmark Experiments}
\label{sec:exp}
This section delineates the experimental results to assess the robustness of BEV algorithms under varied corrupted scenarios. Subsection~\ref{sec:setup} details the experimental setup and methodology. Subsection~\ref{sec:camera-only} discusses the performance of camera-only BEV models on the \textit{nuScenes-C} dataset, highlighting the impact of different corruptions on model robustness. Subsection~\ref{sec:fusion} explores the resilience of camera-LiDAR fusion models to sensor-specific corruptions and complete sensor failures. Finally, Subsection~\ref{sec:validity} evaluates the realism and validity of the synthetic corruption used in the experiments.

\begin{figure*}[ht]
    \centering
    \begin{minipage}[t]{0.31\textwidth}
        \centering
        \subfloat[mCE \textit{vs.} NDS]{
            \label{fig:nds-mce}
            \includegraphics[width=\linewidth]{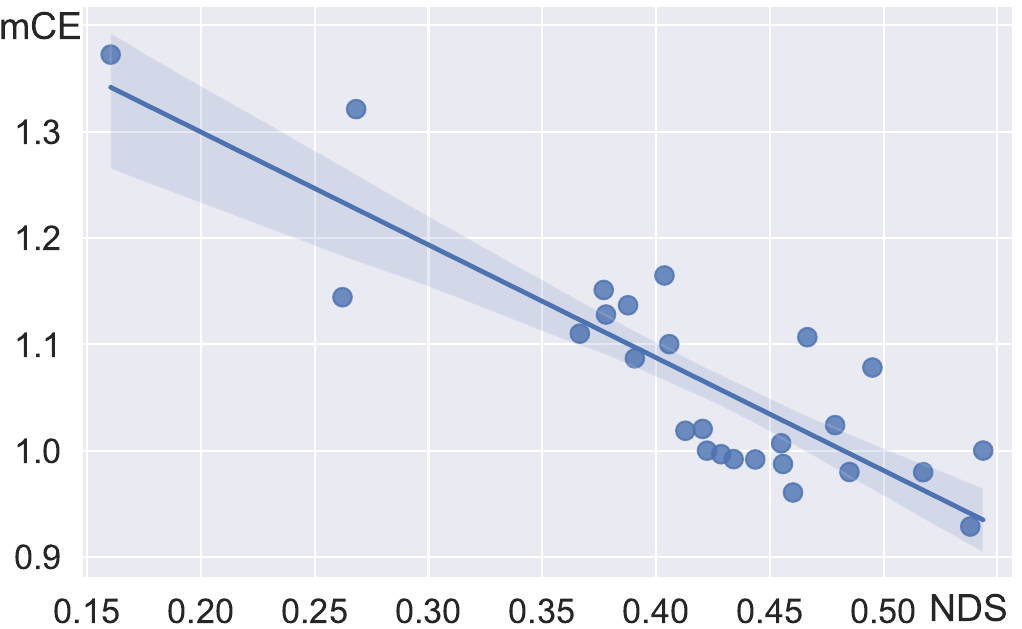}
        }
    \end{minipage}~~
    \begin{minipage}[t]{0.31\textwidth}
        \centering
        \subfloat[mRR \textit{vs.} NDS]{
            \label{fig:nds-mrr}
            \includegraphics[width=\linewidth]{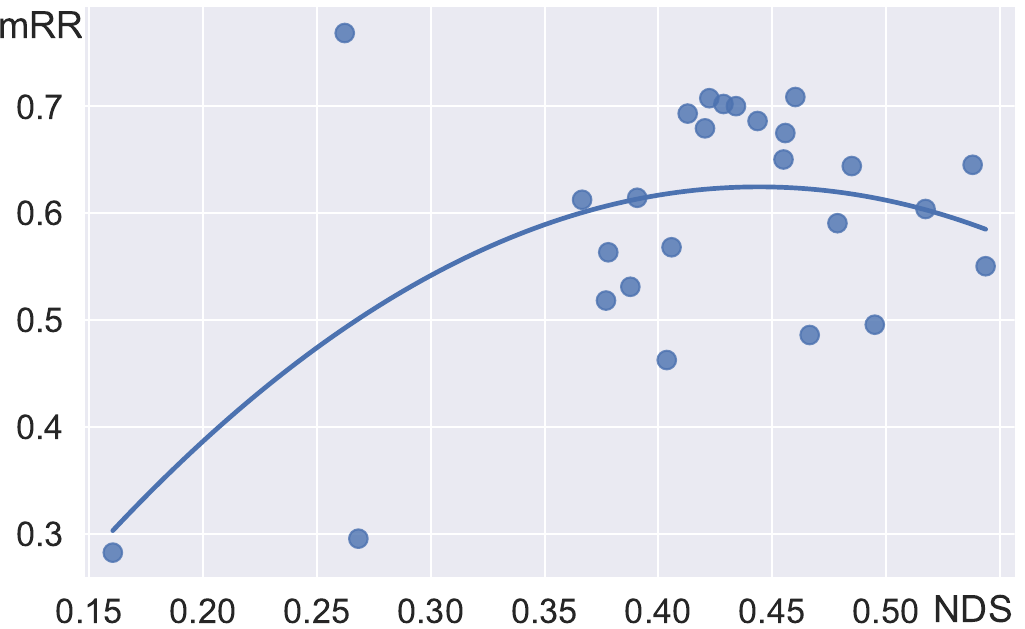}
        }
    \end{minipage}~~
    \begin{minipage}[t]{0.32\textwidth}
        \centering
        \subfloat[Depth Estimation Error \textit{vs.} RR.]{
            \label{fig:bevdepth-depth-error}
            \includegraphics[width=\linewidth]{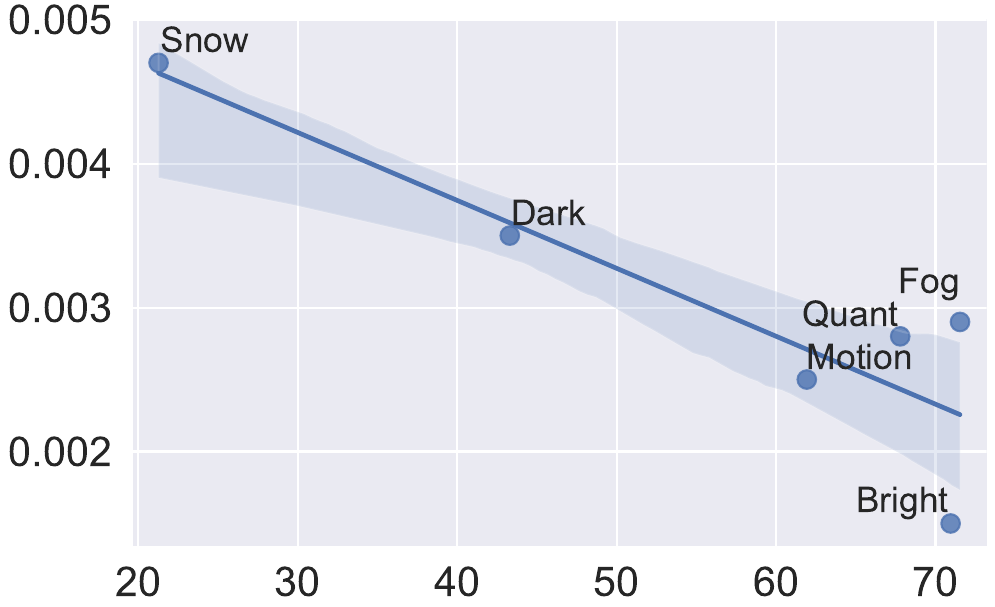}
        }
    \end{minipage}
    \caption{(a): The mCE metric shows a linear relationship with ``clean'' performance while (b): the mRR metric confronts the risk of decreasing. (c): We observe strong correlations where large depth estimation errors under \textit{Snow} and \textit{Dark} tend to cause drastic performance drops.}
    \label{fig:overall}
\end{figure*}

\subsection{Experimental Settings}
\label{sec:setup}
In our study, we use the official model configurations and public checkpoints provided by open-sourced codebases, whenever applicable; we also train additional model variants with minimal modifications to conduct experiments under controlled settings. To facilitate access to all model checkpoints and configurations, we have compiled a ``model zoo'', which can be accessed through our repository\footnote{The RoboBEV model zoo is publicly accessible at \url{https://github.com/Daniel-xsy/RoboBEV/blob/master/zoo/README.md}.}.

We re-implemented several models, including BEVDet (r101) \cite{huang2021bevdet}, PolarFormer (vov) \cite{jiang2022polarformer}, and SRCN3D (vov) \cite{shi2022srcn3d}, tailored to our investigative requirements. For BEVDet (r101), to ensure fair comparisons, we choose to preserve the input resolution consistent with BEVDet (r50). For the original versions of both PolarFormer (vov) and SRCN3D (vov), the models were initialized using checkpoints from DD3D~\cite{park2021pseudo}, which had previously trained on the nuScenes \textit{trainval} set, causing information leakage. To mitigate this and ensure fair comparisons, we re-implemented the two models, initiating them via the FCOS3D~\cite{wang2021fcos3d} model, without further alterations. Specifically, the VoVNet-V2 iterations \cite{lee2020centermask} of the FCOS3D models were first trained for depth estimation on the DDAD15M dataset~\cite{guizilini20203d} and then underwent fine-tuning on the nuScenes training set.

Furthermore, for a comprehensive overview, metrics for each corruption type were deduced by averaging results across all three severity levels. In our study, DETR3D~\cite{wang2022detr3d} was chosen as the baseline for the mCE metric. Our research methodology and the corresponding code were constructed upon the MMDetection3D codebase \cite{mmdet3d,mmdet3d-lidarseg}.

\subsection{Camera-Only Benchmark}
\label{sec:camera-only}
We undertook an exhaustive benchmark analysis of 30 contemporary BEV models on the \textit{nuScenes-C} dataset. The primary outcomes of our investigations are encapsulated in Table~\ref{tab:robodet_model}. Our analysis revealed that all models exhibit a decline in performance across the corrupted dataset.

\begin{table*}[t]
    \centering
    \caption{Benchmark results of perception tasks including Map Segmentation (MS), Depth Estimation (DE), and Semantic Occupancy Prediction (SOP).}
    \vspace{-0.2cm}
    \label{tab:appe-map-seg}
    \scalebox{1}{
    \begin{tabular}{c|r|r|c|cccccccc}
    \toprule
    \textbf{Tasks} & \textbf{Model} & \textbf{Metric} & \textbf{Clean} & \textbf{Camera} & \textbf{Frame} & \textbf{Quant} & \textbf{Motion} & \textbf{Bright} & \textbf{Dark} & \textbf{Fog} & \textbf{Snow} 
    \\\midrule\midrule
    \rowcolor{gray!10}MS & CVT\cite{zhou2022cross} & IoU $\uparrow$ & 0.348 & 0.200 & 0.170 & 0.294 & 0.281 & 0.275 & 0.200 & 0.247 & 0.177
    \\
    \midrule
    DE & SurroundDepth~\cite{wei2023surrounddepth} & Abs Rel $\downarrow$ & 0.280 & 0.485 & 0.497 & 0.334 & 0.338 & 0.339 & 0.354 & 0.320 & 0.423 
    \\\midrule
    \rowcolor{gray!10}SOP & TPVFormer~\cite{huang2023tri} & mIoU $\uparrow$ & 0.521 & 0.274 & 0.229 & 0.381 & 0.386 & 0.490 & 0.373 & 0.466 & 0.193 
    \\
    SOP & SurroundOcc~\cite{wei2023surroundocc} & SC IoU $\uparrow$ & 0.314 & 0.199 & 0.181 & 0.258 & 0.225 & 0.307 & 0.248 & 0.296 & 0.183 
    \\\bottomrule
    \end{tabular}
    }
\end{table*}

\subsubsection{3D Object Detection}
A notable trend emerges when examining the absolute performances on both the nuScenes-C and its ``clean'' counterpart. Specifically, BEV detectors exhibiting superb performance on the standard dataset also tend to showcase commendable performance when faced with out-of-distribution datasets, a trend visually represented in Figure~\ref{fig:nds-mce}. Nevertheless, looking more closely at these results reveals a more complex story. Detectors, despite parallel performance on the ``clean'' dataset, display varied robustness when confronted with diverse corruption types. For example, BEVerse (swin-s) \cite{zhang2022beverse} shows strong resilience during a \textit{Camera Crash}, and PETR (vov) \cite{liu2022petr} performs well in \textit{Snow} conditions. However, both perform poorly in \textit{Dark} settings.

Our investigations further highlight a potential vulnerability in resilience rates across various corruptions. Even though the mCE metric displays a linear correlation between the nuScenes and \textit{nuScenes-C} datasets, the mRR metric elucidates notable disparities among models with comparable baseline performance as shown in Table~\ref{tab:robodet_rr}. This suggests potential overfitting of some models to the nuScenes dataset, thereby compromising their adaptability to the \textit{nuScenes-C} dataset. For instance, despite Sparse4D~\cite{lin2022sparse4d} outperforming DETR3D~\cite{wang2022detr3d} on the ``clean'' dataset, it falls short in terms of mRR metrics across all corruption categories. Moreover, DETR3D performs exceptionally well in \textit{Dark} conditions, in stark contrast to BEVerse (swin-t) \cite{zhang2022beverse}. Despite its strong performance in clean conditions, BEVerse (swin-t) achieves only 12\% relative performance in the dark. Therefore, a comprehensive assessment of cutting-edge models is essential for a complete evaluation of their capabilities.

As shown in Table~\ref{tab:robodet_model}, to gain deeper insights into the model's robustness, we dissected BEV models based on components like training strategies (\textit{e.g.}, pre-training and CBGS~\cite{zhu2019class} resampling), model architectures (\textit{e.g.}, backbone), and learning techniques (\textit{e.g.}, temporal cue learning) and discuss them consequentially in Section~\ref{sec:dis}.

\begin{table*}[t]
    \centering
    \caption{The BEV detection results (NDS) of fusion-based models under different input modalities. Since \textit{Fog} and \textit{Snow} can also affect the LiDAR sensors, these two types of corruptions are not considered for the image-LiDAR-fusion models.}
    \vspace{-0.2cm}
    \label{tab:multimodal}
    \scalebox{1}{
    \footnotesize
    \begin{tabular}{r|c|c|c|cccccccc}
    \toprule
    \textbf{Model} & \textbf{Camera} & \textbf{LiDAR} & \textbf{Clean} & \textbf{Camera} & \textbf{Frame} & \textbf{Quant} & \textbf{Motion} & \textbf{Bright} & \textbf{Dark} & \textbf{Fog} & \textbf{Snow} 
    \\
    \midrule\midrule
    \rowcolor{gray!10}BEVFusion~\cite{liu2022bevfusion} & \checkmark &  & 0.4121 & 0.2777 & 0.2255 & 0.2763 & 0.2788 & 0.2902 & 0.1076 & 0.3041 & 0.1461 
    \\
    BEVFusion~\cite{liu2022bevfusion} & & \checkmark & 0.6928 & $-$ & $-$ & $-$ & $-$ & $-$ & $-$ & $-$ & $-$ 
    \\\midrule
    \rowcolor{gray!10}BEVFusion~\cite{liu2022bevfusion} & \checkmark & \checkmark & 0.7138 & 0.6963 & 0.6931 & 0.7044 & 0.6977 & 0.7018 & 0.6787 & $-$ & $-$
    \\
    TransFusion~\cite{bai2022transfusion} & \checkmark & \checkmark & 0.6887 & 0.6843 & 0.6447 & 0.6819 & 0.6749 & 0.6843 & 0.6663 & $-$ & $-$ 
    \\
    \rowcolor{gray!10}AutoAlignV2~\cite{chen2022autoalignv2} & \checkmark & \checkmark & 0.6139 & 0.5849 & 0.5832 & 0.6006 & 0.5901 & 0.6076 & 0.5770 & $-$ & $-$ 
    \\\bottomrule
    \end{tabular}
    }
\end{table*}

\begin{table*}[t]
    \centering
    \caption{Benchmark results for BEV detection robustness under sensor failures. The models are trained using multi-modal (camera and LiDAR) inputs while tested using single sensor inputs. For LiDAR failure, we keep the partial field of view in front of the ego vehicle since none of the candidate models can even operate with a complete LiDAR modality missing.}
    \vspace{-0.2cm}
    \label{tab:sensor-fail}
    \scalebox{1}{
    \footnotesize
    \begin{tabular}{r|c|c|c|c|c|ccccc}
    \toprule
    \textbf{Model} & \textbf{Train} & \textbf{Camera} & \textbf{LiDAR} & \textbf{NDS} $\uparrow$ & \textbf{mAP} $\uparrow$ & \textbf{mATE} & \textbf{mASE} & \textbf{mAOE} & \textbf{mAVE} & \textbf{mAAE} \\
    \midrule\midrule
    \rowcolor{gray!10}BEVFusion~\cite{liu2022bevfusion} & C & \checkmark &  &  0.4122 & 0.3556 & 0.6677 & 0.2727 & 0.5612 & 0.8954 & 0.2593 
    \\
    BEVFusion~\cite{liu2022bevfusion} & L &  & \checkmark &  0.6927 & 0.6468 & 0.2912 & 0.2530 & 0.3142 & 0.2627 & 0.1858 
    \\\midrule
    \rowcolor{gray!10}BEVFusion~\cite{liu2022bevfusion} & C+L & \checkmark & \checkmark & 0.7138 & 0.6852 & 0.2874 & 0.2539 & 0.3044 & 0.2554 & 0.1874 
    \\
    BEVFusion~\cite{liu2022bevfusion} & C+L & \checkmark &  &  0.3340 {($\downarrow$ 0.3798)} & 0.0789 {($\downarrow$ 0.6063)} & 0.5044 & 0.3073 & 0.4999 & 0.5098 & 0.2338 \\
    \rowcolor{gray!10}BEVFusion~\cite{liu2022bevfusion} & C+L & & \checkmark & 0.6802 {($\downarrow$ 0.0605)} & 0.6247 {($\downarrow$ 0.0605)} & 0.2948 & 0.2590 & 0.3137 & 0.2697 & 0.1844 
    \\\midrule
    TransFusion~\cite{bai2022transfusion} & C+L & \checkmark & \checkmark & 0.6887 & 0.6453 & 0.2995 & 0.2552 & 0.3209 & 0.2765 & 0.1877
    \\
    \rowcolor{gray!10}TransFusion~\cite{bai2022transfusion} & C+L & \checkmark & & 0.3470 {($\downarrow$ 0.3417)} & 0.0343 {($\downarrow$ 0.6110)} & 0.4087 & 0.3091 & 0.4446 & 0.3104 & 0.2282 
    \\
    TransFusion~\cite{bai2022transfusion} & C+L &  & \checkmark & 0.6464 {($\downarrow$ 0.0423)} & 0.5764 {($\downarrow$ 0.0689)} & 0.3171 & 0.2761 & 0.3227 & 0.3124 & 0.1897 
    \\\midrule
    \rowcolor{gray!10}AutoAlignV2~\cite{chen2022autoalignv2} & C+L & \checkmark & \checkmark & 0.6139 & 0.5649 & 0.3300 & 0.2699 & 0.4226 & 0.4644 & 0.1983 \\
    AutoAlignV2~\cite{chen2022autoalignv2} & C+L &  & \checkmark & 0.5651 {($\downarrow$ 0.0448)} & 0.4794 {($\downarrow$ 0.0855)} & 0.3463 & 0.2734 & 0.4361 & 0.4894 & 0.2007 \\
    \bottomrule
    \end{tabular}
    }
\end{table*}

\begin{table*}[t]
    \centering
    \caption{The Resilience Rate (RR) of each BEV detector in our RoboBEV benchmark. Bold: best within the corruptions. \underline{Underlined}: Best across corruptions. Symbol $^\dag$ distinguishes the pre-training version BEVDet. The RR scores are in percentage (\%) and the higher the better.}
    \label{tab:robodet_rr}
    \vspace{-0.2cm}
    \scalebox{0.98}{
    \footnotesize
    \begin{tabular}{r|p{1.1cm}<{\centering}|p{1.1cm}<{\centering}|p{1.cm}<{\centering}p{1.cm}<{\centering}p{1.cm}<{\centering}p{1.cm}<{\centering}p{1.cm}<{\centering}p{1.cm}<{\centering}p{1.cm}<{\centering}p{1.cm}<{\centering}}
    \toprule
    \textbf{Model} & \textbf{NDS} $\uparrow$ & \textbf{mRR} $\uparrow$ & \textbf{Camera} & \textbf{Frame} & \textbf{Quant} & \textbf{Motion} & \textbf{Bright} & \textbf{Dark} & \textbf{Fog} & \textbf{Snow} \\
    \midrule\midrule
    \rowcolor{gray!10}DETR3D~\cite{wang2022detr3d} & 0.4224  & 70.77  & 67.68 & 61.65 & 75.21 & 63.00 & 94.74 & \underline{\textbf{65.96}} & \underline{\textbf{92.61}} & 45.29 
    \\
    DETR3D$_{\text{CBGS}}$~\cite{wang2022detr3d} & 0.4341 & 70.02  & \textbf{68.90} & 61.85 & 74.52 & 58.56 & \underline{\textbf{95.69}} & 63.72 & \underline{\textbf{92.61}} & 44.34 
    \\
    \rowcolor{gray!10}BEVFormer~{\scriptsize (small)}~\cite{li2022bevformer} & 0.4787 & 59.07  & 57.89 & 51.37 & 68.41 & 53.69 & 78.15 & 50.41 & 74.85 & 37.79  
    \\
    BEVFormer~{\scriptsize (base)}~\cite{li2022bevformer} & \textbf{0.5174} & 60.40 & 60.96 & 58.31 & 67.82 & 52.09 & 80.87 & 48.61 & 78.64 & 35.89
    \\
    \rowcolor{gray!10}PETR~{\scriptsize (r50)}~\cite{liu2022petr} & 0.3665 & 61.26  & 63.30 & 59.10 & 67.45 & 62.73 & 77.52 & 42.86 & 78.47 & 38.66
    \\
    PETR~{\scriptsize (vov)}~\cite{liu2022petr} & 0.4550 & 65.03 & 64.26 & 61.36 & 65.23 & 54.73 & 84.79 & 50.66 & 81.38 & \textbf{57.85} 
    \\
    \rowcolor{gray!10}ORA3D~\cite{roh2022ora3d} & 0.4436 & 68.63 & 68.87 & \textbf{61.99} & 75.74 & 59.67 & 91.86 & 58.90 & 89.25 & 42.79
    \\
    PolarFormer~{\scriptsize (r101)}~\cite{jiang2022polarformer} & 0.4602 & \underline{\textbf{70.88}}  & 68.08 & 61.02 & \textbf{76.25} & \textbf{69.99} & 93.52 & 55.50 & \underline{\textbf{92.61}} & 50.07
    \\
    \rowcolor{gray!10}PolarFormer~{\scriptsize (vov)}~\cite{jiang2022polarformer} & 0.4558 & 67.51  & 68.78 & 61.67 & 67.49 & 51.43 & 93.90 & 53.55 & 89.10 & 54.15
    \\\midrule
    SRCN3D~{\scriptsize (r101)}~\cite{shi2022srcn3d} & 0.4286 & \textbf{70.23} & \textbf{68.76} & \textbf{62.55} & \textbf{77.41} & \textbf{60.87} & \textbf{95.05} & \textbf{60.43} & \textbf{91.93} & 44.80
    \\
    \rowcolor{gray!10}SRCN3D~{\scriptsize (vov)}~\cite{shi2022srcn3d} & 0.4205 & 67.95 & 68.37 & 61.33 & 67.23 & 50.96 & 92.41 & 54.08 & 89.75 & \underline{\textbf{59.43}}
    \\
    Sparse4D~{\scriptsize (r101)}~\cite{lin2022sparse4d} & \textbf{0.5438} & 55.04 & 52.83 & 48.01 & 60.87 & 46.23 & 73.26 & 46.16 & 71.42 & 41.54 
    \\
    \midrule
    \rowcolor{gray!10}BEVDet~{\scriptsize (r50)}~\cite{huang2021bevdet} & 0.3770 & 51.83 & 65.94 & 51.03 & 63.87 & 54.67 & 68.04 & 29.23 & 65.28 & 16.58
    \\
    BEVDet~{\scriptsize (tiny)}~\cite{huang2021bevdet} & 0.4037 & 46.26  & 64.63 & 52.39 & 56.43 & 52.71 & 54.27 & 12.14 & 60.69 & 16.84 \\
    \rowcolor{gray!10}BEVDet~{\scriptsize (r101)}~\cite{huang2021bevdet} & $0.3877$ & $53.12$  & $67.63$ & $53.26$ & $65.67$ & $58.42$ & $65.88$ & $28.84$ & $64.35$ & $20.89$ \\
    BEVDet$^\dag$~{\scriptsize (r101)}~\cite{huang2021bevdet}  & $0.3780$ & $56.35$  & $64.60$ & $51.90$ & \underline{\textbf{80.45}} & $68.52$ & $68.76$ & $36.85$ & $54.84$ & $24.84$ \\
    \rowcolor{gray!10}BEVDepth~{\scriptsize (r50)}~\cite{li2022bevdepth}  & $0.4058$ & $56.82$  & $65.01$ & $52.76$ & $67.79$ & $61.93$ & $70.95$ & $43.30$ & $71.54$ & $21.27$ \\
    BEVerse~{\scriptsize (swin-t)}~\cite{zhang2022beverse} & $0.4665$ & $48.60$  & $68.19$ & \underline{\textbf{65.10}} & $55.73$ & $56.74$ & $56.93$ & $12.71$ & $59.61$ & $13.80$ \\
    \rowcolor{gray!10}BEVerse~{\scriptsize (swin-s)}~\cite{zhang2022beverse} & $0.4951$ & $49.57$  & $67.95$ & $50.19$ & $56.70$ & $53.16$ & $68.55$ & $22.58$ & $57.54$ & $19.89$ \\
    SOLOFusion~{\scriptsize (short)}~\cite{Park2022TimeWT} & $0.3907$ & $61.45$  & $65.04$ & $56.18$ & $71.77$ & $66.62$ & $75.92$ & $52.03$ & $76.73$ & $27.28$ \\
    \rowcolor{gray!10}SOLOFusion~{\scriptsize (long)}~\cite{Park2022TimeWT} & $0.4850$ & $64.42$  & $65.13$ & $51.34$ & $74.19$ & \underline{\textbf{71.34}} & \textbf{82.52} & \textbf{58.02} & $82.29$ & \textbf{30.52} \\
    SOLOFusion~{\scriptsize (fusion)}~\cite{Park2022TimeWT} & \textbf{0.5381} & \textbf{64.53}  & \underline{\textbf{70.73}} & $64.37$ & $75.41$ & $67.68$ & $80.45$ & $48.80$ & \textbf{83.26} & $25.57$ \\
    \bottomrule
    \end{tabular}
    }
\end{table*}

\subsubsection{Other BEV Perception Tasks}
Our inquiry also extended to associated tasks, including BEV-centric map segmentation, depth estimation, and occupancy prediction, with outcomes presented in Table~\ref{tab:appe-map-seg}. Adhering to setting 1 from \cite{zhou2022cross}, we reported the Intersection over Union (IoU) for vehicle map-view segmentation results. For depth estimation, we employed the Absolute Relative Difference (Abs Rel) score, and for semantic occupancy prediction, we used the mean Intersection over Union (mIoU). For comprehensive metric definitions, readers can consult the original publications \cite{zhou2022cross, wei2023surrounddepth, huang2023tri, wei2023surroundocc}. These results, spanning diverse perception tasks, offer an enriched perspective on BEV model capabilities and constraints.
It is worth noting that many BEV-centric perception models struggle with specific corruptions, such as \textit{Dark} and \textit{Snow}. This reveals a common vulnerability among BEV models, compromising their reliability in real-world scenarios.

\begin{table*}[t]
    \centering
    \caption{Validity study of using corruptions as data augmentations during the model training to improve the cross-domain robustness of BEV detectors.}
    \vspace{-0.2cm}
    \label{tab:validity}
    \scalebox{0.96}{
    \footnotesize
    \begin{tabular}{r|c|c|c|c|c|ccccc}
    \toprule
    \textbf{Model} & \textbf{Training Set} & \textbf{Testing Set} & \textbf{CorruptAug} & \textbf{NDS} $\uparrow$ & \textbf{mAP} $\uparrow$ & \textbf{mATE} & \textbf{mASE} & \textbf{mAOE} & \textbf{mAVE} & \textbf{mAAE} \\
    \midrule\midrule
    \rowcolor{gray!10}DETR~\cite{wang2022detr3d} & nuScenes train & nuScenes val & & 0.4224 & 0.3468 & 0.7647 & 0.2678 & 0.3917 & 0.8754 & 0.2108 
    \\
    DETR~\cite{wang2022detr3d} & nuScenes train & nuScenes val & \checkmark & 0.4242 {($\uparrow$ 0.0018)} & 0.3511 & 0.7655 & 0.2736 & 0.4130 & 0.8487 & 0.2119 
    \\\midrule
    \rowcolor{gray!10}FCOS3D~\cite{wang2021fcos3d} & day train & day val &  & 0.3867 & 0.3045 & 0.7651 & 0.2576 & 0.5001 & 1.2102 & 0.1321 
    \\
    FCOS3D~\cite{wang2021fcos3d} & day train & day val & \checkmark & 0.3883 {($\uparrow$ 0.0196)}  & 0.3073 & 0.7630 & 0.2581 & 0.5043 & 1.1782 & 0.1286 
    \\\midrule
    \rowcolor{gray!10}FCOS3D~\cite{wang2021fcos3d} & day train & night val &  & 0.0854 & 0.0162 & 1.0434 & 0.6431 & 0.8241 & 1.8505 & 0.7597 
    \\
    FCOS3D~\cite{wang2021fcos3d} & day train & night val & \checkmark & 0.1245 {($\uparrow$ 0.0391)} & 0.0265 & 1.0419 & 0.4658 & 0.8145 & 2.2727 & 0.6067 
    \\\midrule
    \rowcolor{gray!10}FCOS3D~\cite{wang2021fcos3d} & dry train & dry val &  & 0.3846 & 0.2970 & 0.7744 & 0.2541 & 0.4721 & 1.3199 & 0.1380 
    \\
    FCOS3D~\cite{wang2021fcos3d} & dry train & dry val & \checkmark & 0.3854 {($\uparrow$ 0.0008)} & 0.2992 & 0.7654 & 0.2582 & 0.4824 & 1.3334 & 0.1361 
    \\\midrule
    \rowcolor{gray!10}FCOS3D~\cite{wang2021fcos3d} & dry train & rain val & & 0.3203 & 0.2151 & 0.8994 & 0.2856 & 0.5253 & 1.7129 & 0.1619 
    \\
    FCOS3D~\cite{wang2021fcos3d} & dry train & rain val & \checkmark & 0.3302 {($\uparrow$ 0.0099)} & 0.2266 & 0.8595 & 0.2719 & 0.5559 & 1.5697 & 0.1439 \\
    \bottomrule
    \end{tabular}
    }
\end{table*}

\begin{table*}[t]
\centering
\caption{Comparisons between the baseline models and models with corruption augmentations. The numbers show the nuScenes detection score (NDS).}
\vspace{-0.2cm}
\begin{tabular}[width=\linewidth]{r|c|c|cccccccc|c}
    \toprule
    \textbf{Model} & \textbf{CorruptAug} & \textbf{Clean} & \textbf{Crash} & \textbf{Frame} & \textbf{Quant} & \textbf{Motion} & \textbf{Bright} & \textbf{Dark} & \textbf{Fog} & \textbf{Snow} & \textbf{mRR} $\uparrow$
    \\\midrule\midrule
    \rowcolor{gray!10}BEVFormer \cite{li2022bevformer} &  & {0.5174} & 0.3154 & 0.3017 & 0.3509 & 0.2695 & 0.4184 & 0.2515 & 0.4069 & 0.1857 & 0.6040
    \\
    BEVFormer \cite{li2022bevformer} & \checkmark & {0.5169} & 0.3267 & 0.3069 & 0.4102 & 0.4028 & 0.4418 & 0.3642 & 0.4379 & 0.3808 & 0.7427
    \\\midrule
    \rowcolor{gray!10}DETR3D \cite{wang2022detr3d} &  & {0.4224} & 0.2859 & 0.2604 & 0.3177 & 0.2661 & 0.4002 & 0.2786 & 0.3912 & 0.1913 & 0.7077
    \\
    DETR3D \cite{wang2022detr3d} & \checkmark & {0.4252} & 0.2956 & 0.2462 & 0.3923 & 0.3850 & 0.4159 & 0.3622 & 0.4135 & 0.3827 & 0.8506
    \\\midrule
    \rowcolor{gray!10}PETR \cite{liu2022petr} &  & {0.4550} & 0.2924 & 0.2792 & 0.2968 & 0.2490 & 0.3858 & 0.2305 & 0.3703 & 0.2632 & 0.6503
    \\
    PETR \cite{liu2022petr} & \checkmark & {0.4260} & 0.2807 & 0.2567 & 0.3947 & 0.4003 & 0.4138 & 0.3709 & 0.4137 & 0.3846 & 0.8555
    \\\midrule
    \rowcolor{gray!10}PETRv2 \cite{liu2022petrv2} &  & {0.4860} & 0.4016 & 0.3928 & 0.4418 & 0.4053 & 0.4597 & 0.4129 & 0.4467 & 0.3994 & 0.8642
    \\
    PETRv2 \cite{liu2022petrv2} & \checkmark & {0.4924} & 0.4114 & 0.4029 & 0.4633 & 0.4656 & 0.4734 & 0.4541 & 0.4712 & 0.4601 & 0.9144
    \\\midrule
    \rowcolor{gray!10}BEVDet \cite{huang2021bevdet} & & {0.3500} & 0.2313 & 0.1948 & 0.2435 & 0.2148 & 0.2599 & 0.1408 & 0.2572 & 0.0969 & 0.5854
    \\
    BEVDet \cite{huang2021bevdet} & \checkmark & {0.3326} & 0.2274 & 0.1891 & 0.3033 & 0.3117 & 0.3186 & 0.2600 & 0.3083 & 0.2661 & 0.8210
    \\ 
    \bottomrule
\end{tabular}
\label{tab:corrupt-aug-improve}
\end{table*}

\subsection{Camera-LiDAR Fusion Benchmark}
\label{sec:fusion}
\subsubsection{Camera Sensor Corruptions}
We studied scenarios where cameras are impaired while LiDAR operates optimally, a frequent occurrence in real-world conditions. For instance, LiDAR point cloud capture remains largely unhampered by lighting variations, whereas camera captures can degrade under limited light. Intentionally, we excluded conditions like \textit{Snow} and \textit{Fog}, as they could introduce noise to both camera and LiDAR readings. Results of these studies are depicted in Table~\ref{tab:multimodal}.
Interestingly, multi-modal fusion models maintain high performance even when the camera data is compromised. When provided with original LiDAR and degraded camera inputs, BEVFusion \cite{liu2022bevfusion} consistently outperforms its LiDAR-only counterpart, with a notably higher NDS score of 0.6928, across most types of camera corruptions, except \textit{Dark}. This affirms the efficacy of using LiDAR data even when the camera data is suboptimal.

However, there are circumstances where corrupted camera inputs adversely affect the model's performance. For example, under conditions such as \textit{Camera Crash} and \textit{Motion Blur}, the benefits of incorporating camera features into the model are marginal. Moreover, in the presence of \textit{Dark} corruption, corrupted camera features not only fail to provide useful information but also diminish the efficacy of LiDAR features, leading to a performance drop from an NDS score of 0.6928 to 0.6787. As a result, enhancing the robustness of multi-modal fusion models against input corruption emerges as a crucial avenue for future research.

\subsubsection{Complete Sensor Failure}
Multi-modal fusion models are typically trained using data from both camera and LiDAR sensors. However, the deployed model must function adequately even if one of these sensors fails. We evaluate the performance of our multi-modal model using input from only a single modality, with results presented in Table \ref{tab:sensor-fail}. When simulating camera failure, all pixel values are set to zero. For LiDAR sensor failure, we discovered that no model could perform adequately when all point data are absent (\textit{i.e.}, the NDS falls to zero). Hence, we retain only the points within a $[-$45, 45$]$ degree range in front of the vehicle and discard all others.

Interestingly, our findings indicate that multi-modal models are disproportionately reliant on LiDAR input. In scenarios where LiDAR data is missing, the mAP metrics for BEVFusion~\cite{liu2022bevfusion} and Transfusion~\cite{bai2022transfusion} drop by 89\% and 95\%, respectively. In contrast, the absence of image data leads to a much milder decline in performance. This phenomenon underscores that, during the training phase, point cloud features may disproportionately influence the model, thereby asserting dominance over image-based features in perception tasks.
Such a dependence on LiDAR data introduces a significant vulnerability to multi-modal perception models, particularly because LiDAR sensors are prone to data corruption under adverse weather conditions such as rain, snow, and fog. These observations necessitate further research focused on enhancing the robustness of multi-modal perception systems, especially when one sensory modality is entirely absent.

\subsection{Validity Assessment}
\label{sec:validity}
Since the corruption images are synthesized digitally, it is important to study how closely they are compared to real-world corruption. To study the validity of synthesized images, we conducted two experiments, including the pixel distribution study and corruption-augmented training.

\subsubsection{Pixel Distributions}
Assuming that a corruption simulation is realistic enough to reflect real-world situations, the distribution of a corrupted ``clean'' set should be similar to that of the real-world corruption set. We validate this using ACDC~\cite{sakaridis2021acdc}, nuScenes~\cite{caesar2020nuscenes}, Cityscapes~\cite{cordts2016cityscapes}, and Foggy-Cityscapes~\cite{sakaridis2018semantic}, since these datasets contain real-world corruption data and clean data collected by the same sensor types from the same physical locations. We simulate corruptions using ``clean'' images and compare the distribution patterns with their corresponding real-world corrupted data. We do this to ensure that there is no extra distribution shift from aspects like sensor difference (\textit{e.g.}, FOVs and resolutions) and location discrepancy (\textit{e.g.}, environmental and semantic changes). The pixel histograms are computed jointly on all RGB channels.
As illustrated in the Appendix, the pixel distributions of our synthesized images exhibit similar trends to those of real-world data, thereby showing the potential to use synthesized data to mimic real-world corruptions from a pixel qualitative analysis perspective.

\subsubsection{Corruption-Augmented Training}
\label{sec:corrupt-aug}
Furthermore, if the corruption simulation is realistic enough to reflect real-world situations, a corruption-augmented model should achieve better generalizability than the ``clean'' model when tested on real-world corruption datasets. Also, the corruption-augmented model should also show better performance on the clean dataset. We validate this using nuScenes, nuScenes-Night, and nuScenes-Rain. We adopt FCOS3D~\cite{wang2021fcos3d} as the baseline and train the model with corruption augmentation. For nuScenes-Night and nuScenes-Rain, we train the model on the Day-train and Dry-train splits and evaluate on the Day-val, Night-val, Dry-val, and Rain-val splits. The results can be seen in Table~\ref{tab:validity}. We observe that using synthesized images as the data augmentation strategy successfully improves the cross-domain robustness. Specifically, in day-to-night domain transitions, we observe a significant performance drop from 0.3867 to 0.0854 in the baseline model due to the large domain gap. However, when trained with corruption augmentation, the model's cross-domain performance improves by 45.8\%, thereby validating the validity of our synthesized images.

\subsection{Robustness Improvement}
Building upon our experiment, we further study how to fundamentally improve the robustness of BEV perception models. In Subsection~\ref {sec:corrupt-aug-improve}, we simply use the corruption as data augmentation as in Subsection~\ref{sec:corrupt-aug}. In Subsection~\ref{sec:clip-backbone}, we resort to advanced CLIP~\cite{radford2021learning} as the model backbone since CLIP is well-known to show promising out-of-distribution generality~\cite{miller2021accuracy, nguyen2022quality}.

\begin{figure}
    \centering
    \includegraphics[width=\linewidth]{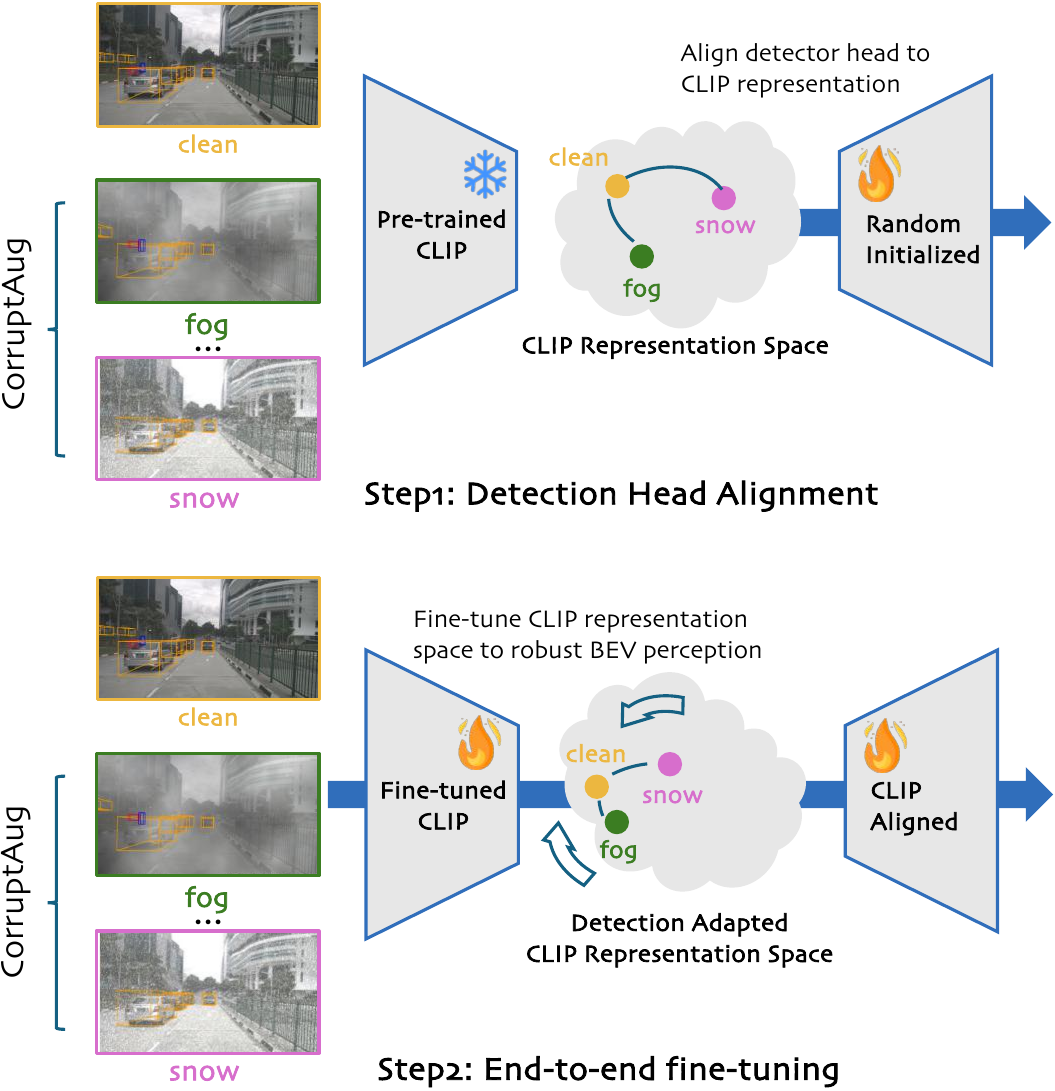}
    \vspace{-0.6cm}
    \caption{Two steps to transfer CLIP~\cite{radford2021learning} robustness to BEVDet~\cite{huang2021bevdet}. The first step is to align the detection head to the frozen CLIP backbone with corruption-augmented inputs. The second step end-to-end fine-tunes the backbone and detection head to enhance the robustness.}
    \label{fig:clip-method}
    \vspace{-0.5cm}
\end{figure}

\begin{figure*}[t]
    \centering
    \includegraphics[width=\linewidth]{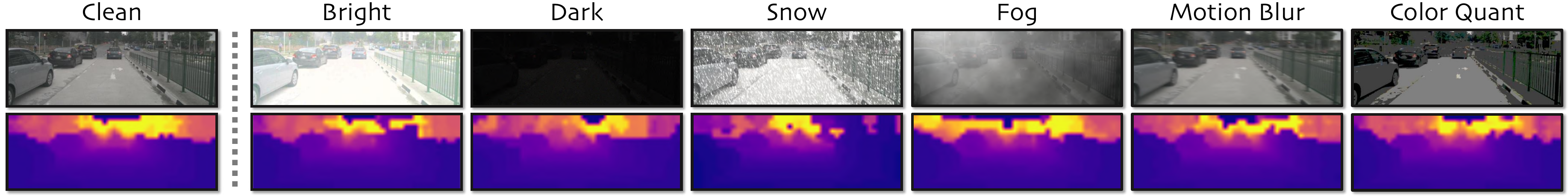}
    \vspace{-0.6cm}
    \caption{Depth estimation results of BEVDepth~\cite{li2022bevdepth} under different corruption types. The results exhibit a different sensitivity for each scenario.}
    \label{fig:depth-estimation}
    \vspace{-0.1cm}
\end{figure*}

\begin{table*}[t]
    \centering
    \caption{We implement the CLIP~\cite{radford2021learning} backbone to improve the robustness of BEVDet~\cite{huang2021bevdet}. We observe that the frozen CLIP backbone cannot transfer well to BEV perception tasks. Head Alignment can further improve the performance while retaining robustness.}
    \vspace{-0.2cm}
    \resizebox{\linewidth}{!}{
        \begin{tabular}{r|c|c|c|c|cccccccc|c}
        \toprule
        \textbf{Backbone} & \textbf{Freeze} & \textbf{Head Align} & \textbf{CorruptAug} & \textbf{Clean} & \textbf{Crash} & \textbf{Frame} & \textbf{Quant} & \textbf{Motion} & \textbf{Bright} & \textbf{Dark} & \textbf{Fog} & \textbf{Snow} & \textbf{mRR} $\uparrow$
        \\\midrule\midrule
        \rowcolor{gray!10}Baseline & - & - &  & 0.3500 & 0.2313 & 0.1948 & 0.2435 & 0.2148 & 0.2599 & 0.1408 & 0.2572 & 0.0969 & 58.54
        \\
        Baseline & - & - & \checkmark & 0.3326 & 0.2274 & 0.1891 & 0.3033 & 0.3117 & 0.3186 & 0.2600 & 0.3083 & 0.2661 & 82.10
        \\\midrule
        \rowcolor{gray!10}CLIP~\cite{radford2021learning} & \checkmark & & & 0.2223 & 0.1544 & 0.1179 & 0.1574 & 0.0816 & 0.1563 & 0.0651 & 0.1659 & 0.0371 & 52.61
        \\
        CLIP~\cite{radford2021learning} & \checkmark & & \checkmark & 0.2093 & 0.1447 & 0.1110 & 0.1840 & 0.1808 & 0.1826 & 0.1127 & 0.1939 & 0.0916 & 71.74
        \\\midrule
        \rowcolor{gray!10}CLIP~\cite{radford2021learning} &  &  &  & 0.3609 & 0.2409 & 0.2090 & 0.2422 & 0.2163 & 0.2804 & 0.1317 & 0.2739 & 0.1119 & 59.10
        \\
        CLIP~\cite{radford2021learning} &  &  & \checkmark & 0.3434 & 0.2330 & 0.2098 & 0.3109 & 0.3217 & 0.3192 & 0.2374 & 0.3154 & 0.2735 & 80.84
        \\\midrule
        \rowcolor{gray!10}CLIP~\cite{radford2021learning} &  & \checkmark &  & \textbf{0.3710} & \textbf{0.2547} & 0.2121 & 0.2562 & 0.2280 & 0.2940 & 0.1508 & 0.2920 & 0.1282 & 61.18
        \\
        CLIP~\cite{radford2021learning} &  & \checkmark & \checkmark & 0.3667 & 0.2514 & \textbf{0.2140} & \textbf{0.3430} & \textbf{0.3484} & \textbf{0.3553} &\textbf{ 0.2922} & \textbf{0.3527} & \textbf{0.3166} & \textbf{84.32}
        \\\bottomrule
        \end{tabular}
    }
    \label{tab:clip}
\end{table*}

\subsubsection{Limitation of Corruption Augmentations}
\label{sec:corrupt-aug-improve}
We start with how to improve the robustness from a data-centric perspective. Specifically, we study if treating synthesized corruption data as a data augmentation alone is enough to solve the problem. To systematically assess the effectiveness of this approach, we apply corruption augmentation to five models within our benchmark, with results detailed in Table~\ref{tab:corrupt-aug-improve}. Our findings indicate that corruption augmentation substantially enhances performance against semantic corruptions within our proposed dataset, particularly for corruption types that originally posed significant challenges to the models (\textit{e.g.}, \textit{Motion Blur}, \textit{Snow}). However, augmenting training data with sensor corruption scenarios, such as the omission of camera information, does not significantly improve robustness. This outcome underscores the limitation of improving robustness solely from a data perspective and highlights the necessity for future research to develop more sophisticated modules capable of handling incomplete input scenarios. However, it should be noted that when using our benchmark as a robustness assessment, synthesized data augmentation should not be a proper practice as it defeats the purpose of the benchmark to analyze robustness from a model perspective.

\subsubsection{Foundation Model Backbone}
\label{sec:clip-backbone}
We further analyze how to improve robustness from both data and model perspectives. Recent studies~\cite{miller2021accuracy, nguyen2022quality} have demonstrated that foundation models, trained on internet-scale datasets in an unsupervised manner, exhibit notable generality compared to models trained on conventional datasets (\textit{e.g.}, ImageNet~\cite{deng2009imagenet}). Inspired by these findings, we explore the potential to transfer the generality of foundation models to BEV perception tasks. Specifically, we investigate three distinct approaches for utilizing the CLIP backbone in our models: (1) freezing the backbone while training only the detector head; (2) fine-tuning both the backbone and the detector head; (3) freezing the backbone and training the detection head, then fine-tuning the whole model together, since previous work~\cite{wortsman2022robust} shows that end-to-end fine-tuning, even though improve in-distribution performance, can compromise robustness on out-of-distribution dataset.

The three methods are shown in Figure~\ref{fig:clip-method} and the results are shown in Table ~\ref{tab:clip}. The first observation is that CLIP is not well optimized for BEV perception tasks, as indicated by the low clean performance when we freeze the CLIP backbone. Additionally, end-to-end fine-tuning with randomly initialized detection heads leads to minimal improvement. Interestingly, when applying corruption augmentation, the CLIP backbone shows little improvement to the baseline model. The improvement in mRR is only 0.56 and, when equipped with corruption augmentation, the mRR is even lower. Finally, our two-stage training can effectively improve the performance while also transferring the robustness of CLIP towards BEV perception tasks, especially with equipped with corruption augmentation. The robustness improvement surpasses the end-to-end fine-tuned CLIP by a notable margin. For instance, the NDS score under \textit{Dark}, \textit{Fog}, and \textit{Snow} improve by 23.1\%, 11.8\%, and 15.8\% respectively. However, even if we show it's possible to improve robustness by leveraging both data and model, the purpose of RoboBEV is mainly to focus on model aspects instead of data. Therefore, it's not a proper practice for future works to evaluate the RoboBEV benchmark with those synthesized corruption as augmentation.

\begin{table}[t]
    \centering
    \caption{Ablation study on model pre-training. Row (a): The PETR~\cite{liu2022petr} baseline. Row (b): The model with the M-BEV~\cite{chen2024m} pre-training. We observe that masked pre-training effectively improves robustness to sensor failure.}
    \vspace{-0.2cm}
    \resizebox{0.99\linewidth}{!}{%
    \begin{tabular}{c|r|cccc}
    \toprule
    \textbf{\#} & \textbf{Model} & \textbf{Crash} & \textbf{Frame} & \textbf{Quant} & \textbf{Motion}
    \\\midrule\midrule
    \rowcolor{gray!10}(a) & PETR~\cite{liu2022petr} & 0.2347 & 0.2336 & 0.2915 & 0.2695
    \\
    (b) & \textit{w/} M-BEV~\cite{chen2024m} & 0.2759 & 0.2715 & 0.3175 & 0.2860
    \\\midrule
    \textbf{\#} & \textbf{Model} & \textbf{Bright} & \textbf{Dark} & \textbf{Fog} & \textbf{Snow}
    \\\midrule\midrule
    \rowcolor{gray!10}(a) & PETR~\cite{liu2022petr} & 0.3029 & 0.2648 & 0.2919 & 0.2463 
    \\
    (b) & \textit{w/} M-BEV~\cite{chen2024m} & 0.3298 & 0.2844 & 0.3191 & 0.2583 
    \\
    \bottomrule
    \end{tabular}
    }
    \label{tab:mbev}
    \vspace{-0.5cm}
\end{table}
\section{Analysis and Discussion}
\label{sec:dis}
This section delves into the nuanced relationships between model designs and robustness under various corruptions. Subsection~\ref{sec:depth} explores the performance of depth estimation models and highlights their susceptibility to image corruptions. Subsection~\ref{sec:pretrain} discusses the impact of pre-training on model resilience across different corruptions. Subsection~\ref{sec:temporal} examines the benefits of temporal fusion. Subsection~\ref{sec:backbone} compares the robustness of different model backbones. Subsection~\ref{sec:corrupt-ana} analyzes how pixel distribution changes correlate with model performance under corruption. Finally, Subsection~\ref{sec:detail-metric} provides detailed metrics on how specific model errors.

\begin{figure*}[ht]
    \centering
    \subfloat[Pre-train - mCE]{
        \label{fig:nds-mce-pretrain}
        \includegraphics[width=0.232\linewidth]{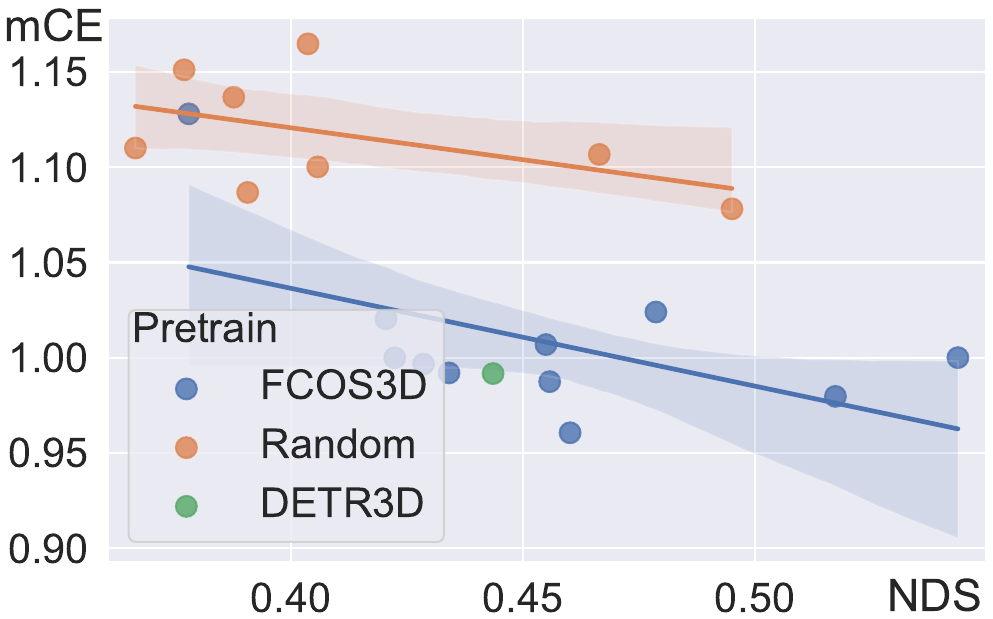}
    }
    ~~
    \subfloat[Pre-train - mRR]{
        \label{fig:nds-mrr-pretrain}
        \includegraphics[width=0.232\linewidth]{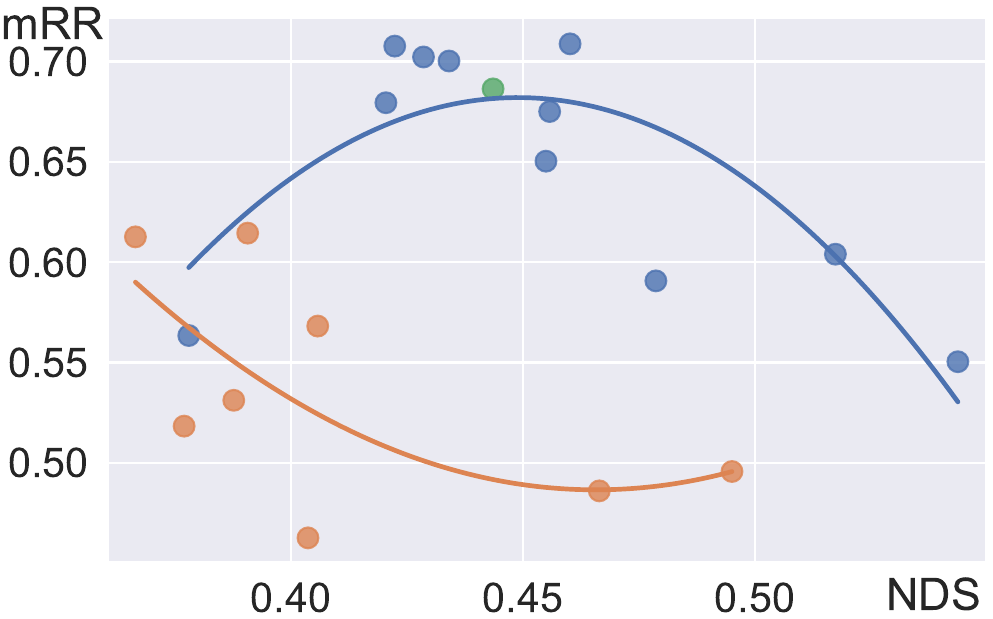}
    }
    ~~
    \subfloat[Depth - mCE]{
        \label{fig:nds-mce-depth}
        \includegraphics[width=0.232\linewidth]{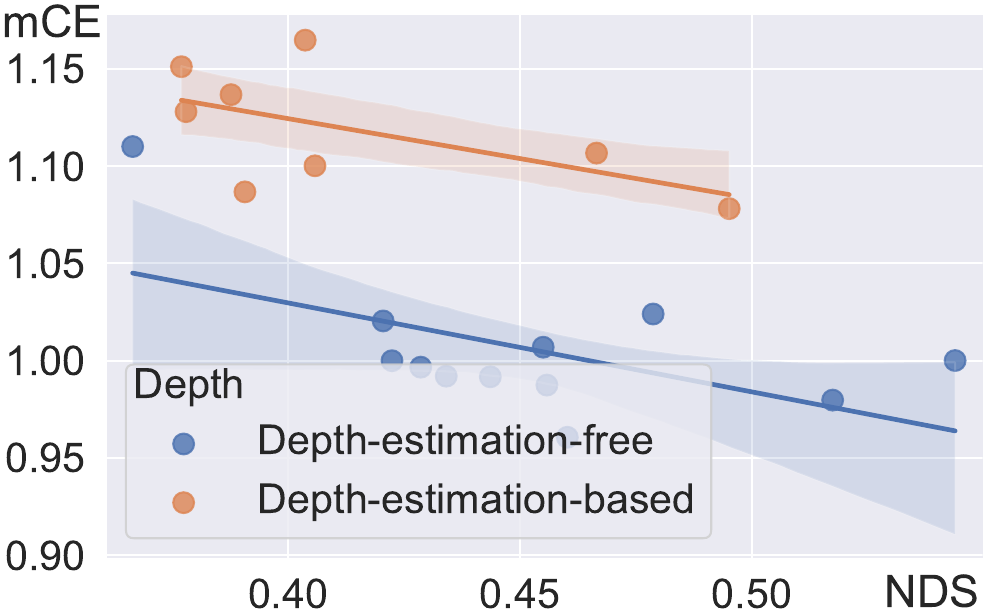}
    }
    ~~
    \subfloat[Depth - mRR]{
        \label{fig:nds-mrr-depth}
        \includegraphics[width=0.232\linewidth]{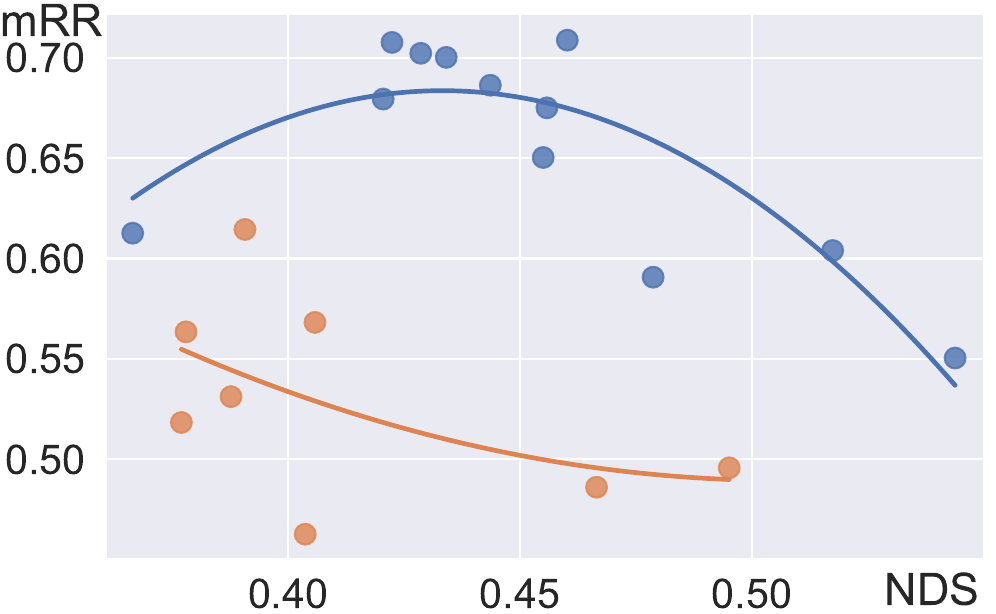}
    }
    \label{fig:pretrain}
    \vspace{-0.1cm}
    \caption{Ablation study. We observe that pre-training strategies together with depth-free bird's eye view transformation provide the models with better robustness. We do not consider SOLOFusion \cite{Park2022TimeWT} long-term fusion here since it utilizes 16 frames, which is much larger than other methods.}
    \vspace{-0.2cm}
\end{figure*}

\begin{figure*}[t]
    \centering
    \subfloat[Camera Crash - CE]{
        \label{fig:nds-mce-cam-temp}
        \includegraphics[width=0.23\linewidth]{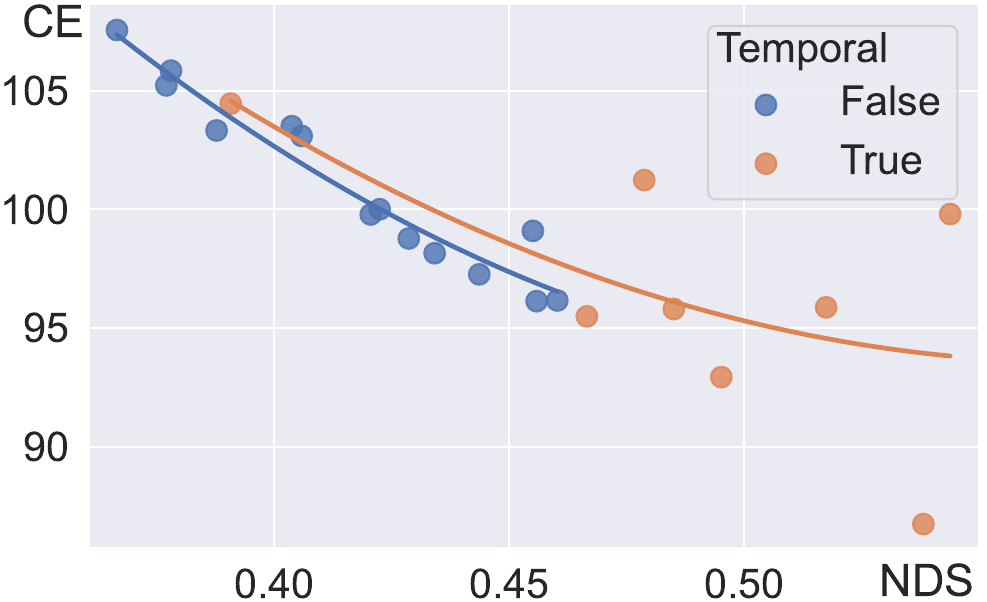}
    }
    ~~
    \subfloat[Camera Crash - RR]{
        \label{fig:nds-mrr-cam-temp}
        \includegraphics[width=0.23\linewidth]{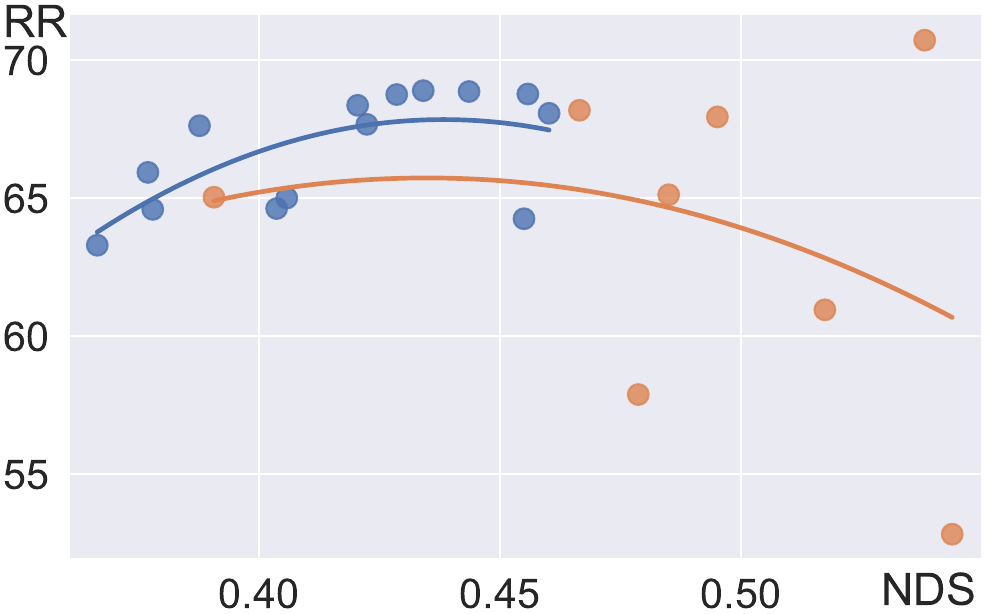}
    }
    ~~
    \subfloat[Frame Lost - CE]{
        \label{fig:nds-mce-frame-temp}
        \includegraphics[width=0.23\linewidth]{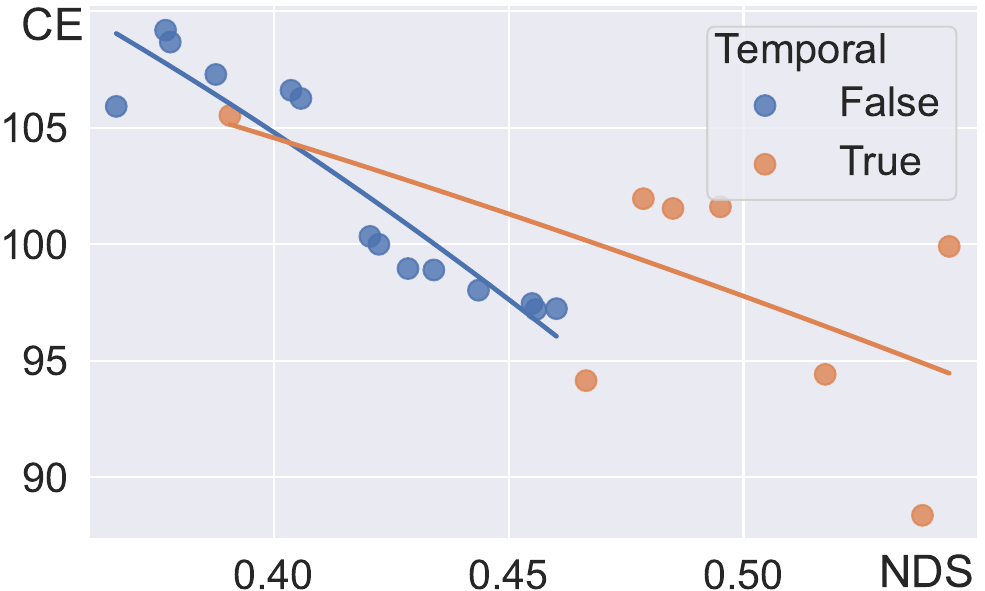}
    }
    ~~
    \subfloat[Frame Lost - RR]{
        \label{fig:nds-mrr-frame-temp}
        \includegraphics[width=0.23\linewidth]{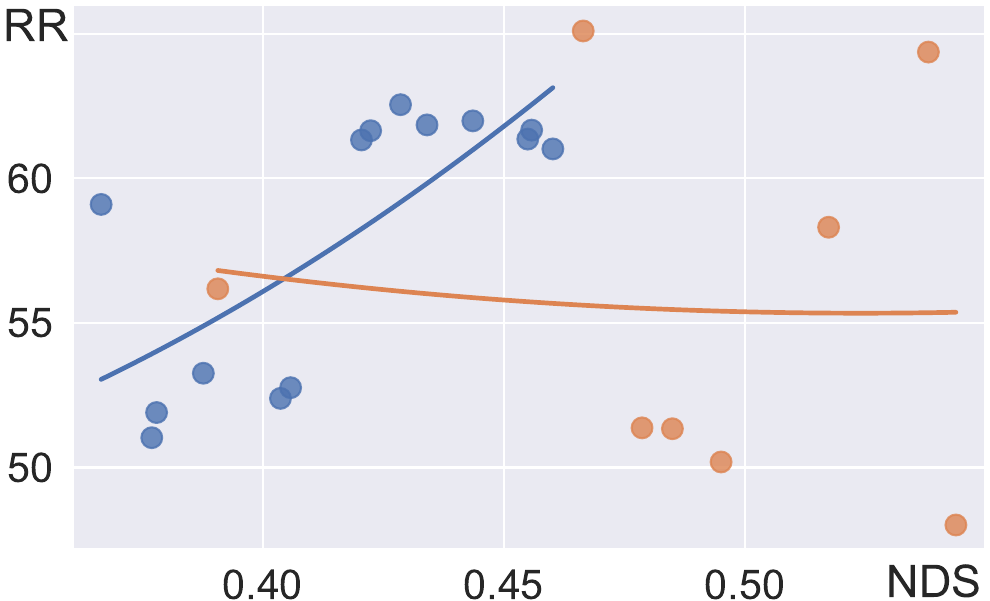}
    }
    \vspace{0.cm}
    \caption{Ablation study. We observe that not all the models with temporal fusion exhibit better robustness under \textit{Camera Crash} and \textit{Frame Lost}. However, they have exhibited certain potential since models with the lowest mCE metric are always those that utilize temporal information.}
    \label{fig:temporal}
\end{figure*}

\begin{figure}
    \centering
    \includegraphics[width=\linewidth]{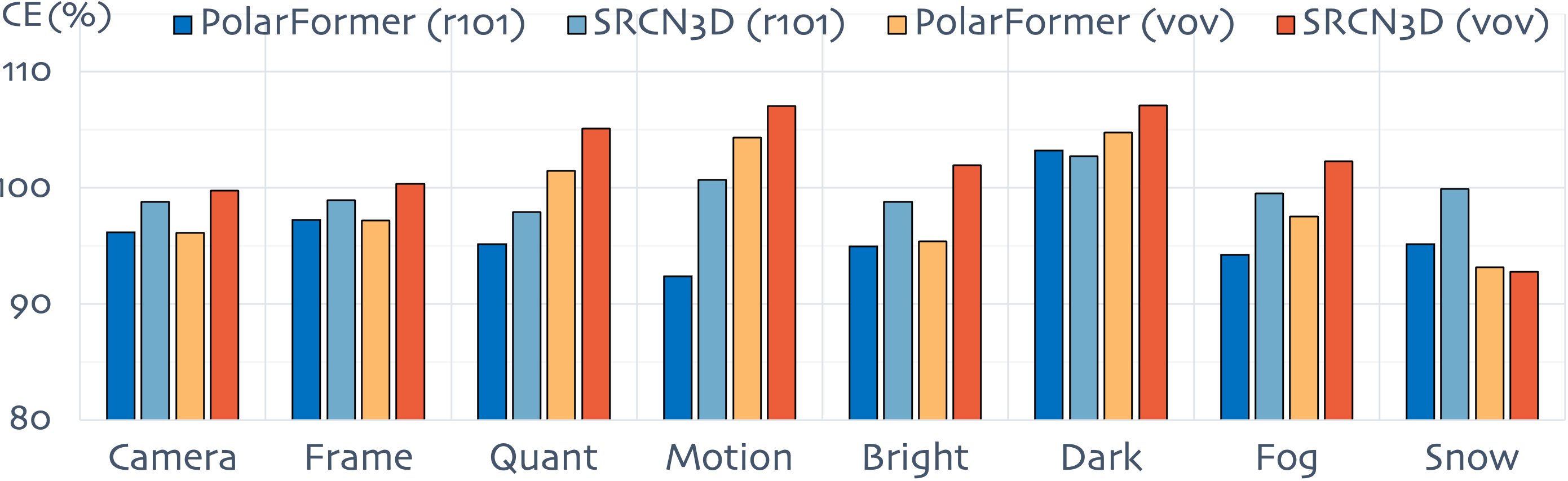}
    \vspace{-0.6cm}
    \caption{Backbone comparison: ResNet \cite{he2016deep} \textit{vs.} VoVNet-V2 \cite{lee2020centermask}. We compare the absolute corruption error (\textit{i.e.}, CE) since the two variants have similar ``clean'' performances.}
    \label{fig:backbone}
\end{figure}

\begin{figure}[t]
    \centering
    \includegraphics[width=\linewidth]{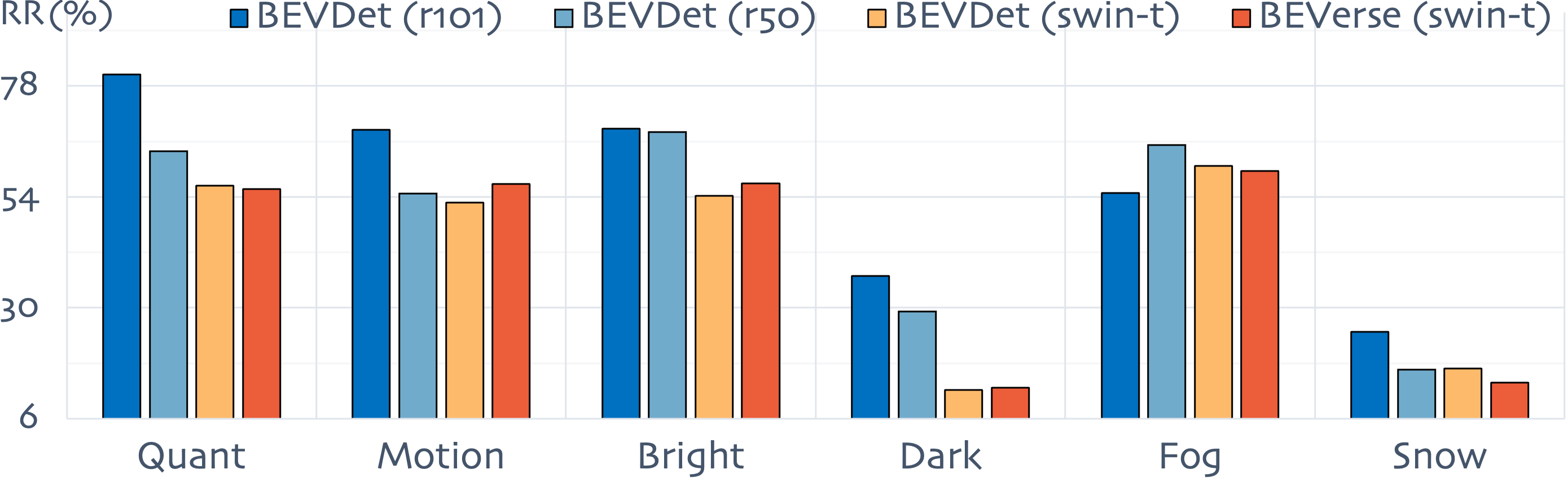}
    \vspace{-0.6cm}
    \caption{Backbone comparison: ResNet \cite{he2016deep} \textit{vs.} Swin Transformer \cite{liu2021swin}. We compare the relative corruption error (\textit{i.e.}, RR) since the two variants have different ``clean'' performances.}
    \label{fig:swin-backbone}
\end{figure}

\begin{figure}[t]
    \centering
    \includegraphics[width=\linewidth]{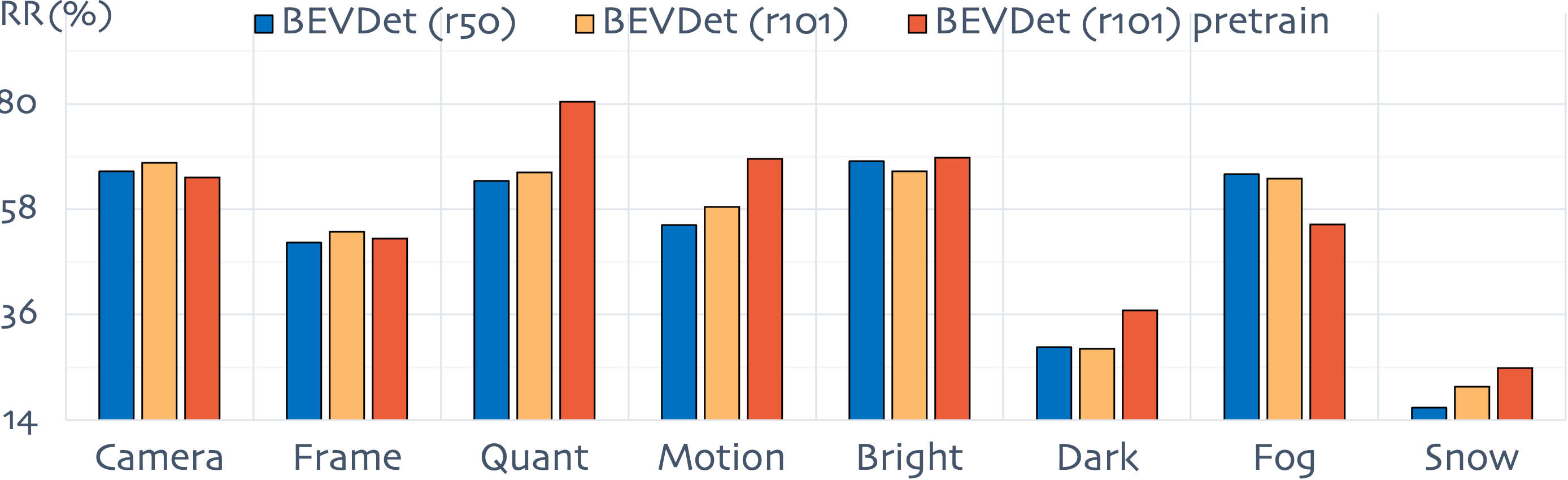}
    \vspace{-0.6cm}
    \caption{Pre-training comparisons. We compare the resilience rate (RR) between BEVDet~\cite{huang2021bevdet} with and without using model pre-training.}
    \label{fig:bevdet-pretrain}
\end{figure}

\subsection{Depth Estimation}
\label{sec:depth}
- \textit{Depth-free BEV transformations show better robustness.}
Our analysis reveals that depth-based approaches suffer from severe performance degradation when exposed to corrupted images as shown in Figure~\ref{fig:nds-mce-depth} and~\ref{fig:nds-mrr-depth}. Moreover, we undertake a comparative study to evaluate the intermediate depth estimation results of BEVDepth~\cite{li2022bevdepth} under corruptions. To this end, we compute the mean square error (MSE) between ``clean'' inputs and corrupted inputs. Our findings indicate an explicit correlation between vulnerability and depth estimation error, as presented in Figure~\ref{fig:bevdepth-depth-error}. Specifically, \textit{Snow} and \textit{Dark} corruptions significantly affect accurate depth estimation, leading to the largest performance drop. These results provide further support for our conclusion that the performance of depth-based approaches can suffer significantly if the depth estimation is not accurate enough. The depth estimation results under corruptions can be seen from Figure~\ref{fig:depth-estimation}, where we can see significant differences under certain corruptions (\textit{e.g.}, \textit{Snow}) compared to ``clean'' inputs.

\begin{figure*}
    \centering
    \subfloat[DETR3D]{
        \includegraphics[width=0.33\linewidth]{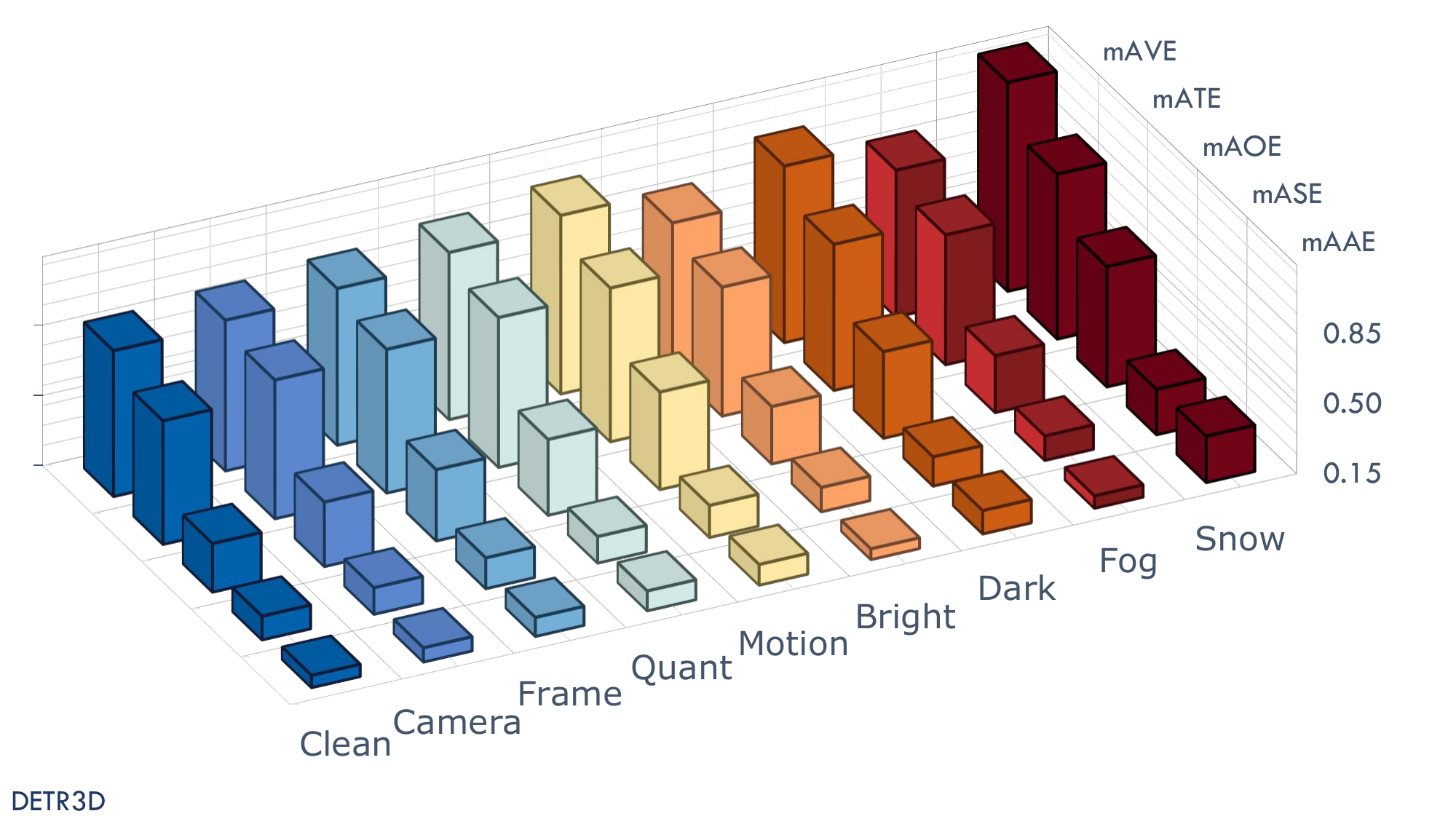}
    }
    \subfloat[BEVFormer]{
        \includegraphics[width=0.33\linewidth]{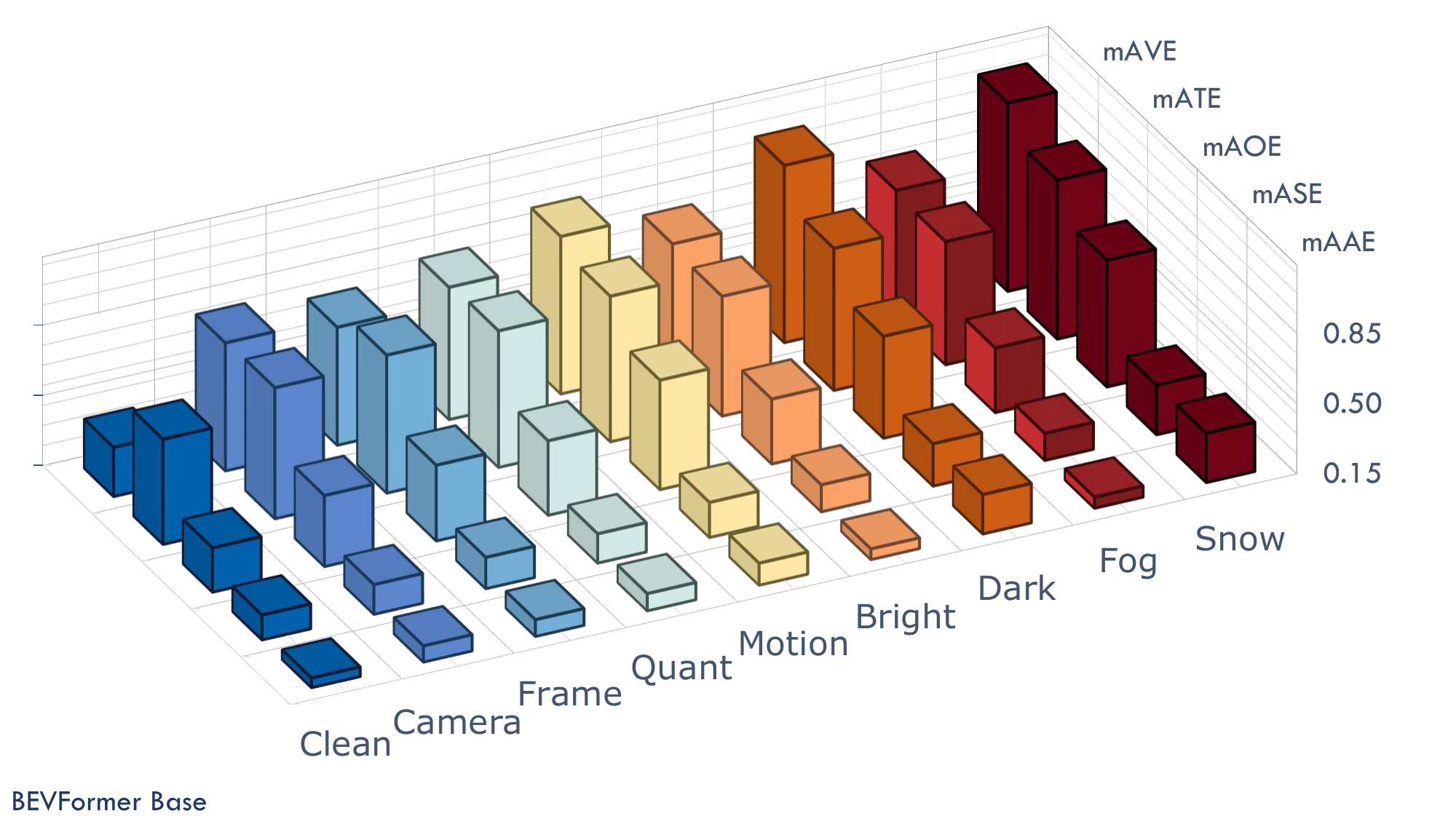}
        \label{fig:3d-bevformer}
    }
    \subfloat[PolarFormer]{
        \includegraphics[width=0.33\linewidth]{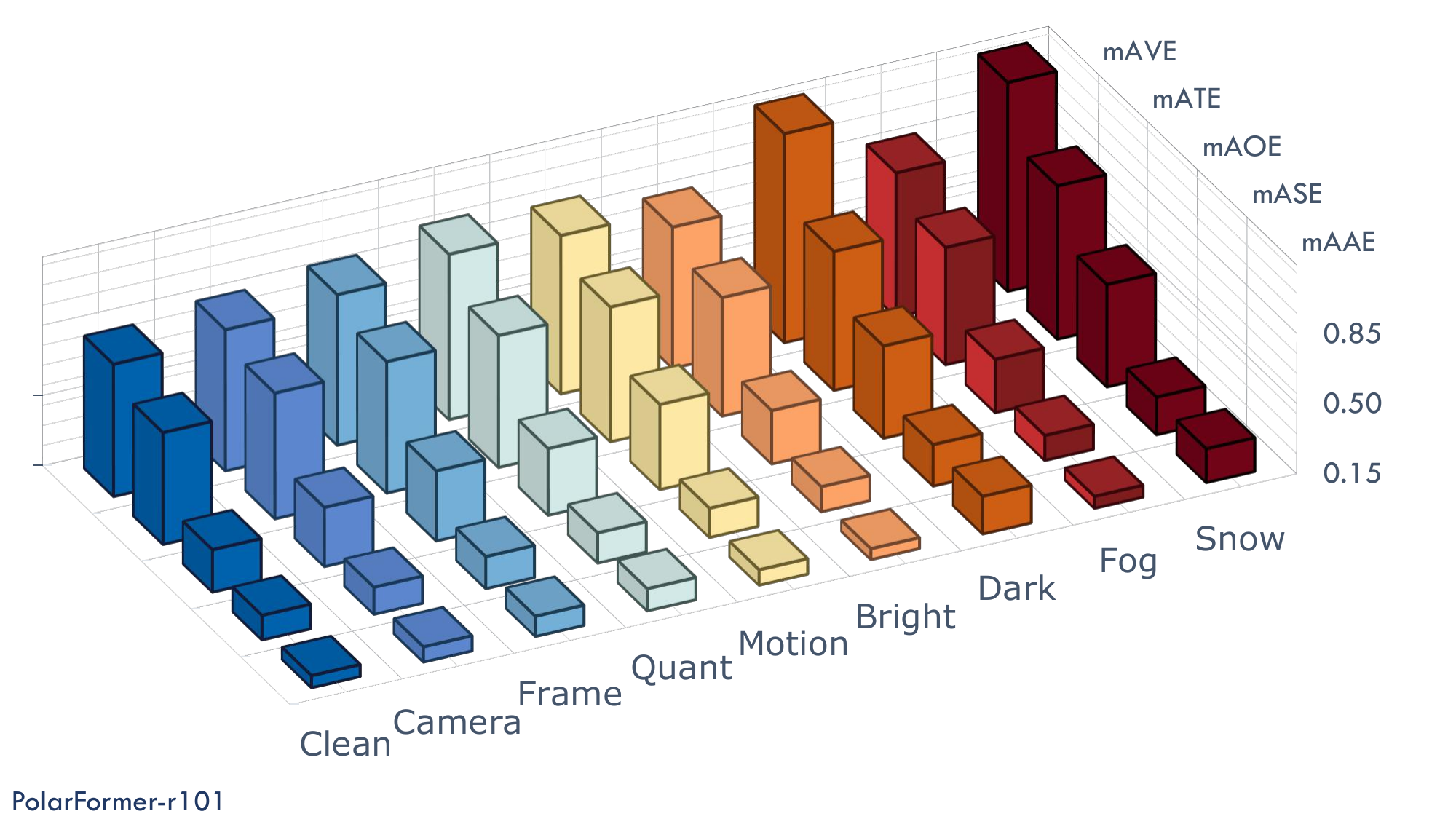}
    }\\
    \subfloat[BEVDet]{
        \includegraphics[width=0.33\linewidth]{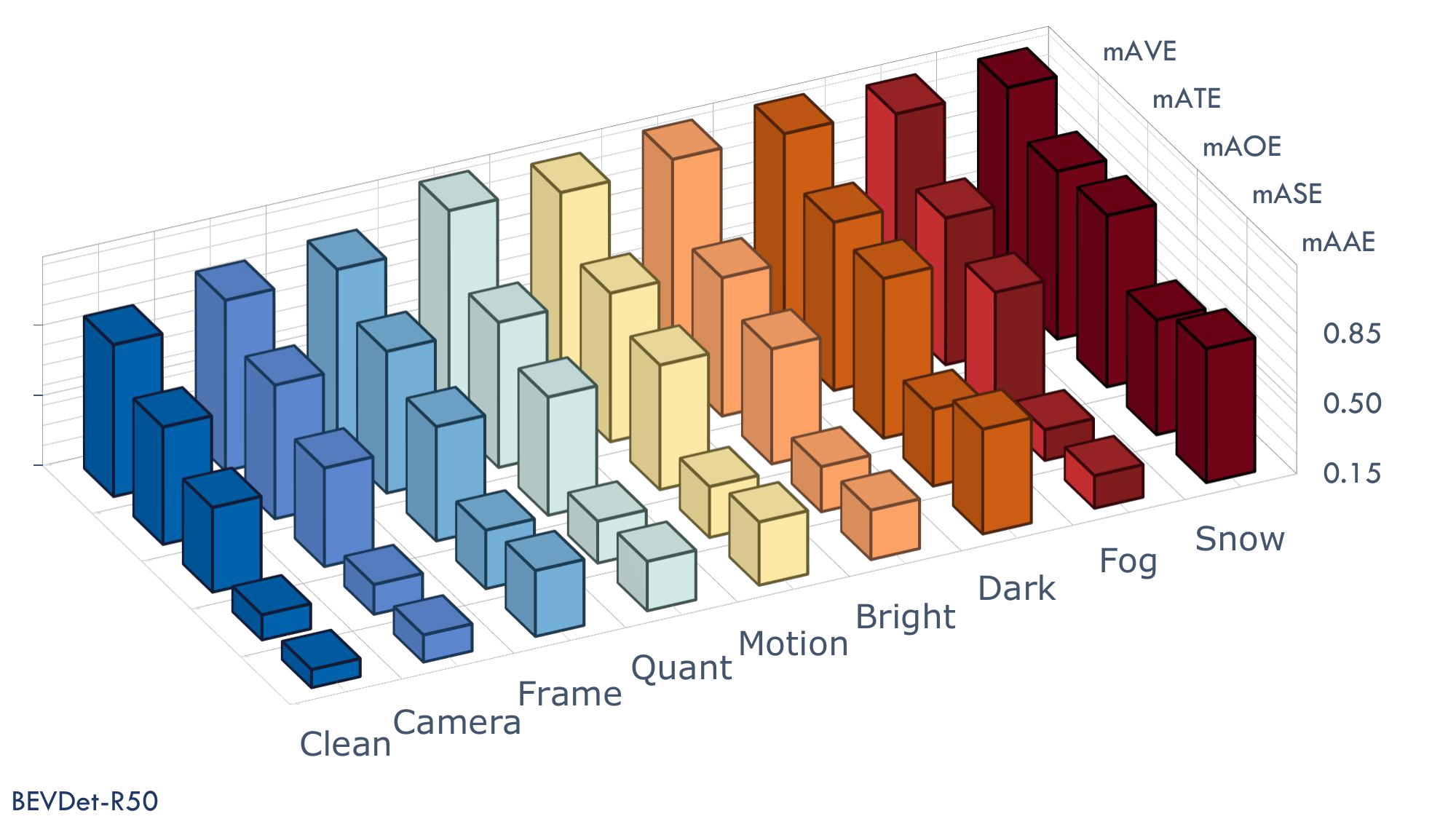}
    }
    \subfloat[BEVDepth]{
        \includegraphics[width=0.33\linewidth]{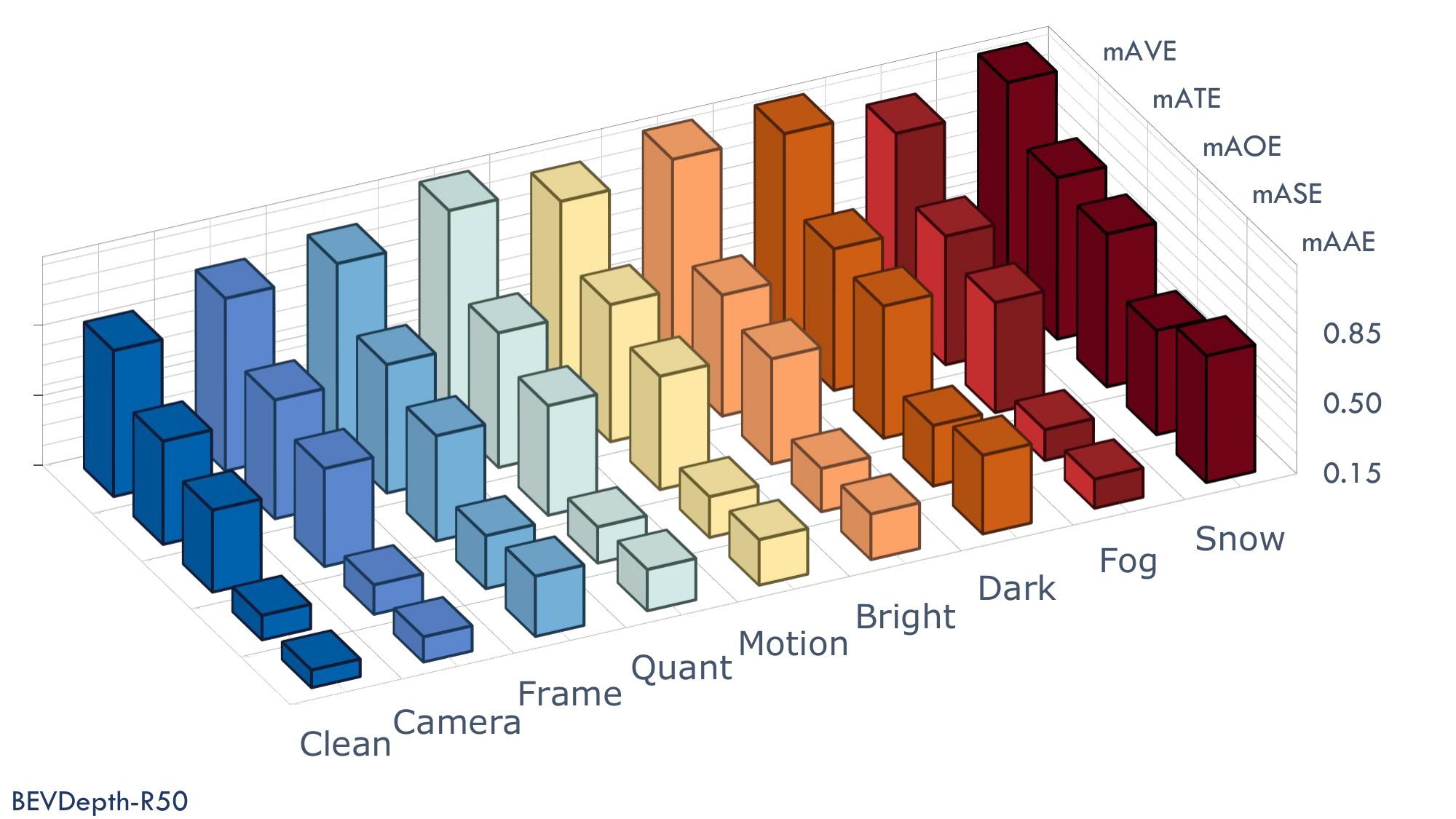}
    }
    \subfloat[BEVerse]{
        \includegraphics[width=0.33\linewidth]{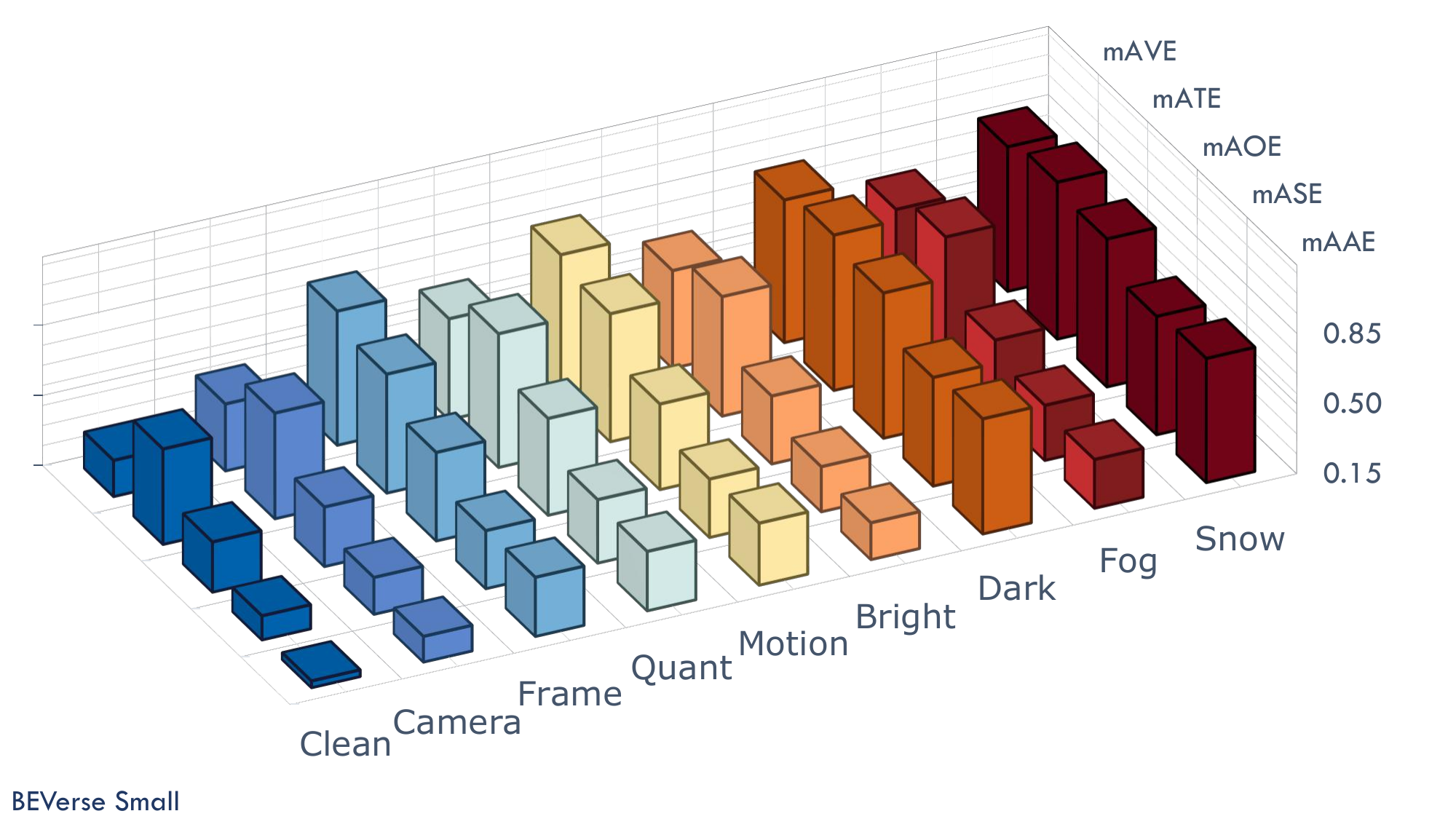}
        \label{fig:3d-beverse}
    }
    \vspace{0.2cm}
    \caption{Benchmark results of task-specific metrics reported on \textit{nuScenes-C} other than NDS, under different corruption types in our benchmark.}
    \label{fig:stat-3d}
\end{figure*}

\subsection{Model Pre-Training}
\label{sec:pretrain}
- \textit{Pre-training improves robustness across a wide range of semantic corruptions while does not help with temporal corruptions.} The effectiveness of these strategies for improving model robustness is illustrated in Figure~\ref{fig:nds-mce-pretrain} and Figure~\ref{fig:nds-mrr-pretrain}, where models that utilize pre-training largely outperform those not. For controlled comparison, we re-implement the BEVDet (r101)~\cite{huang2021bevdet} model using the FCOS3D~\cite{wang2021fcos3d} checkpoint as initialization. Our results, presented in Figure~\ref{fig:bevdet-pretrain}, show that pre-training can significantly improve mRR across a wide range of corruptions (except \textit{Fog}) even if it has lower ``clean'' NDS (0.3780 \textit{vs.} 0.3877). Specifically, under \textit{Color Quant}, \textit{Motion Blur}, and \textit{Dark} corruptions, the mRR metric improves by 22.5\%, 17.2\%, and 27.8\%, respectively. It is worth noting that pre-training mainly improves most semantic corruptions and does not improve temporal corruptions. Even though, the pre-trained BEVDet still largely lags behind those depth-free counterparts. Therefore, we can conclude that pre-training together with the depth-free bird's eye view transformation provides models with strong robustness. More recently, M-BEV~\cite{chen2024m} proposed a masked pre-training task to enhance robustness under incomplete sensor input. We compare a masked pre-trained PETR with the corresponding baseline. The results can be seen in Table~\ref{tab:mbev}. We find M-BEV can most effectively improve robustness towards incomplete sensor output, and reveal the promise of masked image modeling pre-train in BEV perception tasks.

\subsection{Temporal Fusion}
\label{sec:temporal}
- \textit{Temporal fusion has the potential to yield better absolute performance under corruptions. Fusing longer temporal information largely helps with robustness.}
We are particularly interested in examining how models utilizing temporal information perform under temporal corruptions. We find SOLOFusion~\cite{Park2022TimeWT} which fuses wider and richer temporal information performs extremely well compared to its short-only and long-only versions. In terms of \textit{Camera Crash}, the short-only and long-only versions have close resilience rate performance shown in Table~\ref{tab:robodet_rr} (65.04 \textit{vs.} 65.13). However, the fusion version improves to 70.73, which is the highest among all the candidate models. Similarly, the fusion version improves the resilience rate by almost 10\%  compared to the other two versions under \textit{Frame Lost} corruption. Moreover, the RR metric of its long-only version outperforms its short-only counterpart on a wide range of corruption types, which indicates the great potential of utilizing longer temporal information.

\begin{figure}
    \centering
    \includegraphics[width=\linewidth]{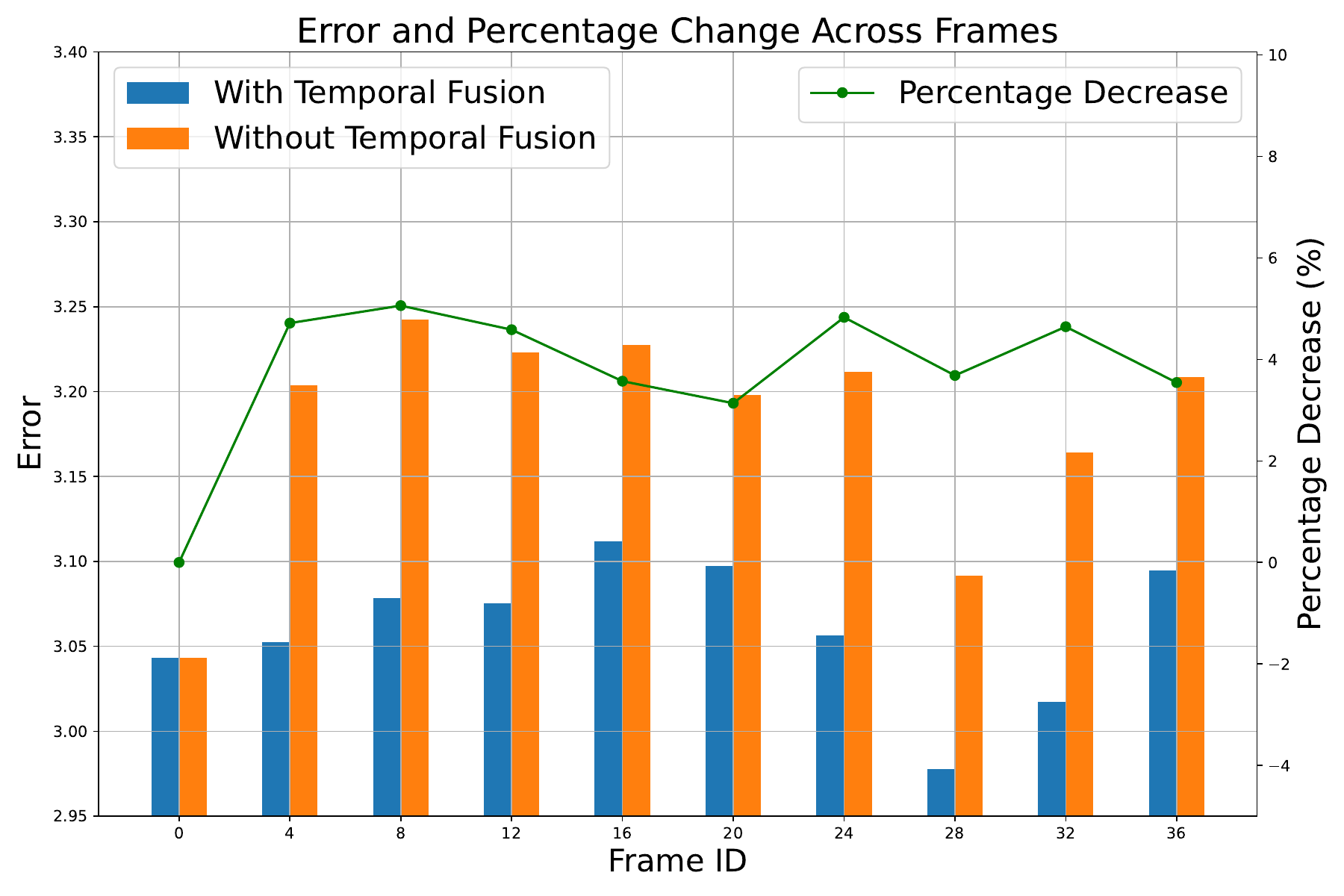}
    \vspace{-0.6cm}
    \caption{Visualized feature errors between the corruption inputs and the ``clean'' inputs. We observe that temporal fusion exhibits a certain potential to mitigate the noises caused by corrupted input images.}
    \label{fig:temp-error}
\end{figure}

To further investigate the impact of temporal fusion on enhancing the corruption robustness, we employ the BEVFormer model to assess feature errors with and without the integration of temporal information compared to ``clean'' temporal inputs. We compute the mean squared error (MSE) between corrupted inputs, both with and without temporal information, and ``clean'' inputs with temporal information. The results are shown in Figure~\ref{fig:temp-error}. We note a trend of increasing error over time in the temporal fusion model (illustrated by the blue bar), attributable to error accumulation under consecutive corrupted inputs. Despite this, temporal fusion consistently demonstrates an error mitigation effect across frames.

However, we find that not all models with temporal fusion exhibit better robustness under \textit{Camera Crash} and \textit{Frame Lost}. The robustness is highly correlated to how to fuse history frames and how many frames are used, which emphasizes the importance of evaluating temporal fusion strategies from wider perspectives. The results can be seen in Figure~\ref{fig:temporal}. Nonetheless, temporal fusion remains a potential method to enhance temporal robustness since the models with the lowest Corruption Error (or the highest Resilience Rate) are consistently those that utilize temporal information.

\begin{figure*}
    \centering
    \includegraphics[width=\linewidth]{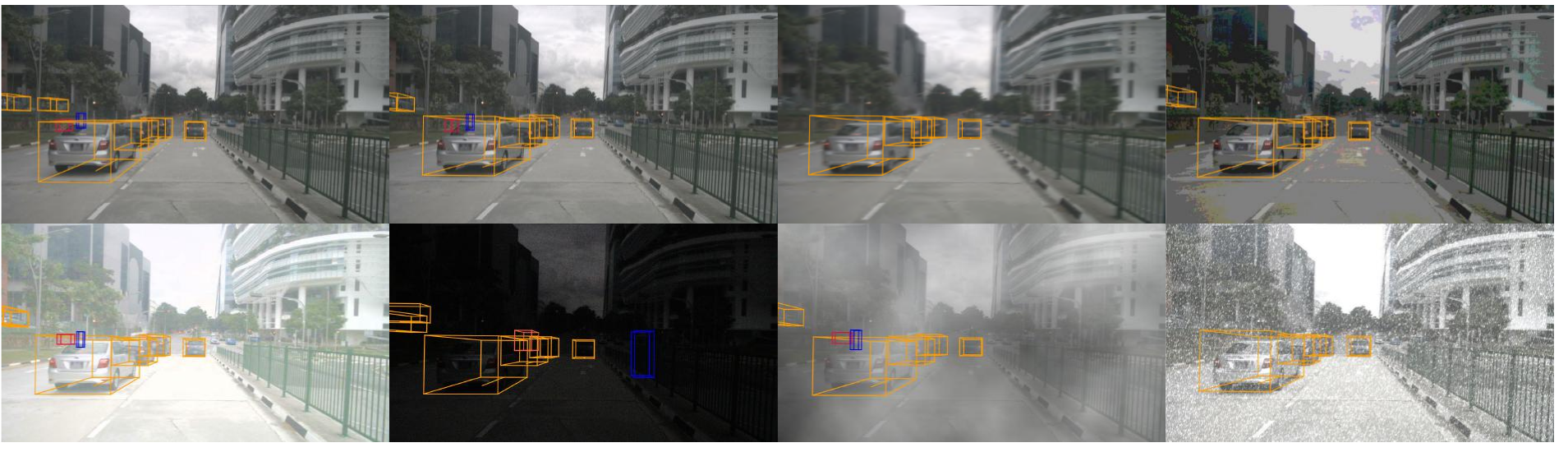}
    \caption{Visualizations of detection results from BEVFormer~\cite{li2022bevformer}. From left to right: top: GT, \textit{Clean}, \textit{Motion}, \textit{Quant}; bottom: \textit{Bright}, \textit{Dark}, \textit{Fog}, \textit{Snow}.}
    \label{fig:visual}
\end{figure*}

\subsection{Backbone}
\label{sec:backbone}
- \textit{The Swin Transformer is more vulnerable towards the lighting changings; VoVNet-V2 is more robust against Snow while ResNet shows better robustness across a wide range of corruptions.} Although ResNet~\cite{he2016deep} and VoVNet~\cite{lee2020centermask} exhibit close standard performance, ResNet-based detectors exhibit consistently superior robustness across a wide range of corruptions, as illustrated in Figure~\ref{fig:backbone}.
Conversely, the VoVNet backbone consistently exhibits better robustness under \textit{Snow} corruptions. Moreover,  Swin Transformer~\cite{liu2021swin} based BEVDet~\cite{huang2021bevdet} demonstrates significant vulnerability towards changes in lighting conditions (\textit{e.g.}, \textit{Bright} and \textit{Dark}). A clear comparison can be found in Figure~\ref{fig:swin-backbone}. Inspired by~\cite{johnson2016perceptual}, we compute the Gramian matrix in the feature space in the features extracted by the backbone model, under clean inputs and corruption inputs. Based on that, we compute the relative error between the ``clean'' and corruption inputs Gramian matrix. The VoVNet has a relative error of 4.17 and 5.43 towards \textit{Snow} and \textit{Motion Blur}, respectively. While ResNet has errors of 11.47 and 2.10. The results align with our end-to-end experiments above, where ResNet is more robust under \textit{Motion Blur} while VoVNet is more robust towards \textit{Snow}. The observed robustness differences between ResNet and VoVNet to \textit{Snow} and \textit{Motion Blur} stem from their architectural designs. ResNet's deep hierarchical features and skip connections make it more sensitive to high-frequency noise like \textit{Snow}, resulting in higher errors. Conversely, VoVNet's dense feature aggregation and efficient feature use enhance robustness to \textit{Snow} by mitigating local noise patterns. However, VoVNet shows higher errors under \textit{Motion Blur}. ResNet's hierarchical feature representation might be better suited to handling the spatial inconsistencies caused by simulated \textit{Motion Blur}. These findings suggest VoVNet excels in spatial noise resilience, while ResNet handles spatial distortions better. Furthermore, the finding can provide insights for feature work to design new loss functions towards aligning the Gramian matrix of features extracted by backbone under different corruptions.

\subsection{Corruptions}
\label{sec:corrupt-ana}
\textit{The relationship between pixel distribution shifts and model performance degradation is not straightforward.}
We calculate the pixel distribution over 300 images sampled from the nuScenes dataset and visualize the pixel histograms shown in Figure~\ref{fig:hist}. Interestingly, the \textit{Motion Blur} causes the least pixel distribution shifts while causing a relatively large performance drop. On the other hand, \textit{Bright} shifts the pixel distribution to higher values, and \textit{Fog} makes fine-grained features more indistinct by shifting the pixel value more agminated. However, these two corruptions only lead to the smallest performance gap, which reveals that model robustness is not simply correlated with pixel distribution.

\subsection{Detailed Metrics}
\label{sec:detail-metric}
- \textit{Velocity prediction errors amplify under corruptions, and attribution and scale errors differ across models.}
While our study predominantly reports the nuScenes Detection Score (NDS) metrics, additional insights into model robustness are illustrated in Figure~\ref{fig:stat-3d}. We find that models incorporating temporal information, such as BEVFormer~\cite{li2022bevformer} and BEVerse~\cite{zhang2022beverse}, exhibit substantially lower mean Absolute Velocity Error (mAVE) compared to those that do not. Nonetheless, even models with temporal fusion are not immune to the adverse effects of image corruption; specifically, velocity prediction errors markedly escalate even under mild illumination alterations. Figure~\ref{fig:3d-bevformer} and \ref{fig:3d-beverse} illustrates that \textit{Motion Blur} corruption detrimentally influences the velocity predictions for both BEVFormer and BEVerse, revealing a significant vulnerability in these models that incorporate temporal data.
Moreover, a closer examination of attribution and scale errors reveals considerable heterogeneity across models. Depth-free models demonstrate a consistent performance in these metrics, while depth-based models display pronounced variability. This observation underscores the heightened susceptibility of depth-based methods to image corruptions and emphasizes the need for further research to enhance their robustness. 
\section{Conclusion}
\label{sec:conclusion}

In this study, we presented the \textit{RoboBEV} benchmark, crafted by incorporating a comprehensive set of eight different natural corruptions to form the \textit{nuScenes-C} dataset. This benchmark serves as a rigorous testing ground for evaluating the out-of-distribution robustness of bird's eye view (BEV) perception models. Additionally, we extended our analyses to account for sensor failures in multi-modal perception frameworks, offering a more holistic view of BEV perception robustness. Through extensive experimentation, we scrutinized various factors influencing the robustness of BEV perception models. Our findings elucidate critical vulnerabilities and strengths across different models and under diverse conditions. Based on our observations, we proposed to leverage the pre-trained CLIP backbone to improve the model's robustness further. By shedding light on these aspects, we aim to furnish the research community with invaluable insights that can guide the development of more robust, future-ready BEV perception systems.

\noindent
{\textbf{Potential Limitation.}
Despite the distinct data corruptions we introduce in our benchmarks, they cannot cover all the out-of-distribution scenarios in real-world applications due to unpredictable complexity. 
Additionally, we mainly analyzed coarse-grained designs among models (\textit{e.g.}, depth estimation) since it is considerably non-trivial to identify the trade-off between fine-grained network designs. Furthermore, we do not include joint corruption on camera and LiDAR, which is essential for some real-world scenarios (\textit{e.g.}, the \textit{Snow} and \textit{Fog} weather cases as shown in Table~\ref{tab:multimodal}).}

\ifCLASSOPTIONcaptionsoff
  \newpage
\fi

\bibliographystyle{IEEEtran}

\end{document}